\documentclass[lettersize,journal]{IEEEtran}
\usepackage{amsmath,amsfonts}
\usepackage{algorithmic}
\usepackage{algorithm}
\usepackage{array}
\usepackage[caption=false,font=normalsize,labelfont=sf,textfont=sf]{subfig}
\usepackage{textcomp}
\usepackage{stfloats}
\usepackage{url}
\usepackage{verbatim}
\usepackage{graphicx}
\usepackage{cite}
\usepackage{mathrsfs}
\usepackage{graphicx}
\usepackage{booktabs}
\usepackage{amsmath}
\usepackage{xcolor}
\usepackage{pifont}
\usepackage[colorlinks, citecolor=blue]{hyperref}
\def\eg{\emph{e.g.}, } 
\def\ie{\emph{i.e.}, }

\begin{document}

\title{Cross-Layer Feature Pyramid Transformer for Small Object Detection in Aerial Images}

\author{Zewen Du, Zhenjiang Hu, Guiyu Zhao, Ying Jin, and Hongbin Ma

\thanks{Zewen~Du, Zhenjiang~Hu, Guiyu~Zhao, Ying~Jin, and Hongbin~Ma are with the National Key Lab of Autonomous Intelligent Unmanned Systems, School of Automation, Beijing Institute of Technology, 100081, Beijing, P. R. China (e-mail: dzw1114@163.com, 3220220769@bit.edu.cn, 3120220906@bit.edu.cn, jinyinghappy@bit.edu.cn, mathmhb@bit.edu.cn).}
}



\maketitle

\begin{abstract}
  Object detection in aerial images has always been a challenging task due to the generally small size of the objects. Most current detectors prioritize the development of new detection frameworks, often overlooking research on fundamental components such as feature pyramid networks. In this paper, we introduce the Cross-Layer Feature Pyramid Transformer (CFPT), a novel upsampler-free feature pyramid network designed specifically for small object detection in aerial images. CFPT incorporates two meticulously designed attention blocks with linear computational complexity: Cross-Layer Channel-Wise Attention (CCA) and Cross-Layer Spatial-Wise Attention (CSA). CCA achieves cross-layer interaction by dividing channel-wise token groups to perceive cross-layer global information along the spatial dimension, while CSA enables cross-layer interaction by dividing spatial-wise token groups to perceive cross-layer global information along the channel dimension. By integrating these modules, CFPT enables efficient cross-layer interaction in a single step, thereby avoiding the semantic gap and information loss associated with element-wise summation and layer-by-layer transmission. In addition, CFPT incorporates global contextual information, which improves detection performance for small objects. To further enhance location awareness during cross-layer interaction, we propose the Cross-Layer Consistent Relative Positional Encoding (CCPE) based on inter-layer mutual receptive fields. We evaluate the effectiveness of CFPT on three challenging object detection datasets in aerial images: VisDrone2019-DET, TinyPerson, and xView. Extensive experiments demonstrate that CFPT outperforms state-of-the-art feature pyramid networks while incurring lower computational costs. The code is available at https://github.com/duzw9311/CFPT.
\end{abstract}


\section{Introduction}

\begin{figure}[t]
  \centering
  \includegraphics[width=0.9\linewidth]{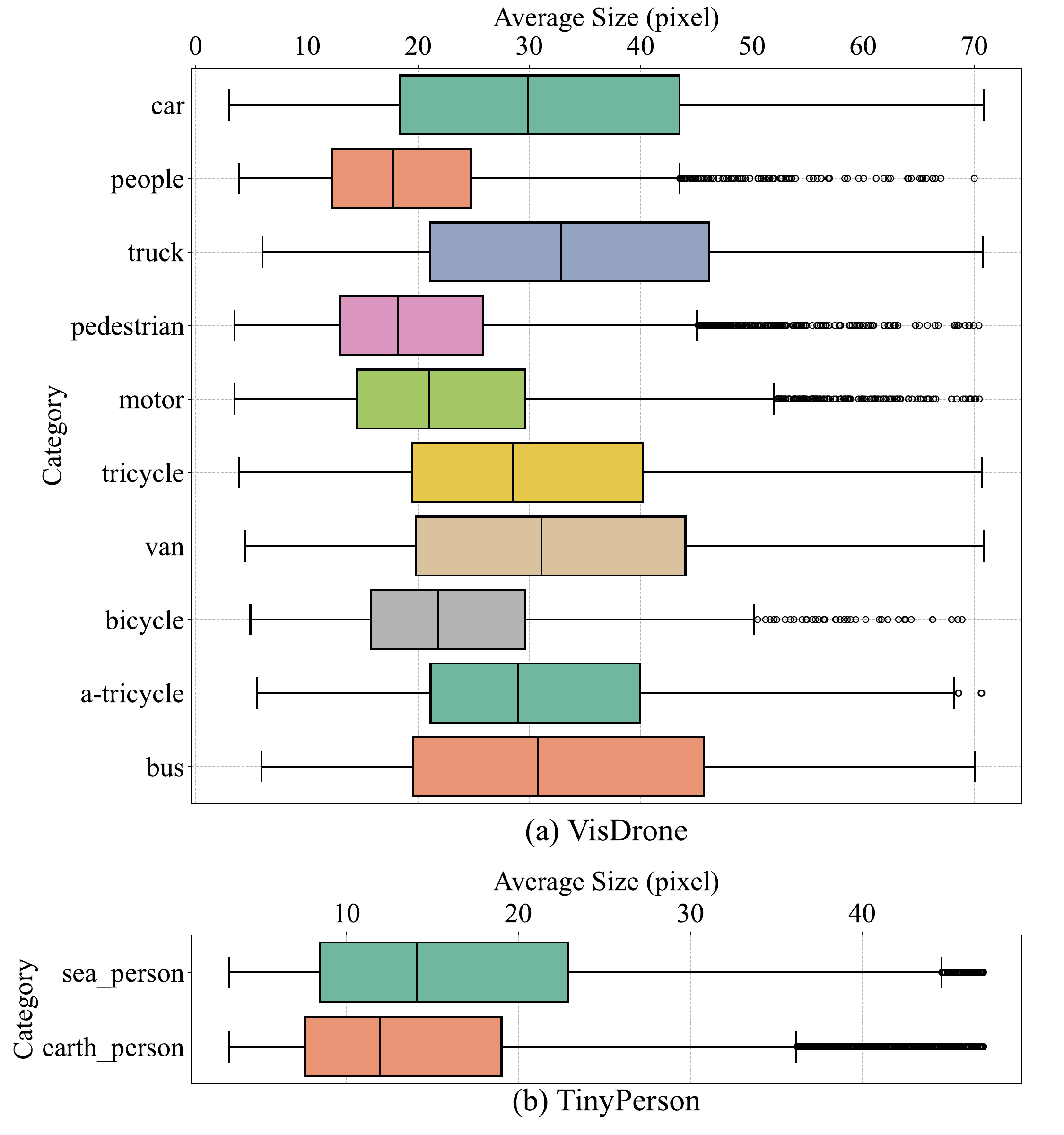}
  \caption{Box plot of scale distribution for (a) VisDrone2019-DET dataset~\cite{du2019visdrone} and (b) TinyPerson dataset~\cite{yu2020scale}. The ordinate represents the category of the annotation bounding boxes, and the abscissa represents the square root of the area of the annotation bounding boxes (\ie $\sqrt{W\times H}$). For clarity, we remove outliers outside the $1.5\times$ Interquartile Range (IQR). }
  \label{fig:dataset_box}
\end{figure}

\IEEEPARstart{B}{enefiting} from advancements in Convolutional Neural Networks (CNNs) and Vision Transformers (ViTs), existing object detectors have achieved significant progress, evolving into essential tools for a wide range of applications, including autonomous driving, face detection, medical image analysis, and industrial quality inspection.

As a subfield of object detection, small object detection still presents greater challenges than traditional object detection tasks, primarily due to the features of small objects being lost or overshadowed by those of larger objects during the convolution and pooling operations. As shown in Fig.\ref{fig:dataset_box}, we present box plots that illustrate the data distribution of two classic small object detection datasets in aerial images: VisDrone2019-DET~\cite{du2019visdrone} and TinyPerson~\cite{yu2020scale}. The box plot shows that the VisDrone2019-DET dataset contains a large number of small objects, with sizes ranging from 20 to 30 pixels, and exhibits significant scale variations across different instances. In contrast, the TinyPerson dataset primarily consists of smaller objects compared to VisDrone2019-DET, with most objects being less than 20 pixels. Additionally, the flying height and shooting angle of the drone significantly impact the scale distribution of objects, leading to relatively lower performance of object detection in aerial images.

To address these challenges, numerous studies have been proposed consecutively. Given the small proportion of foreground objects in drone scenes, existing solutions typically adopt a coarse-to-fine detection scheme~\cite{deng2020global, leng2022pareto, huang2022ufpmp}. In the coarse prediction stage, a standard detector is employed to identify objects and predict dense object clusters. Subsequently, in the refinement stage, these clusters are generally pruned, upsampled, and then re-inputted into the detector for a more precise search. Although such model architectures effectively adapt to drone perspectives and improve the performance of various detectors while maintaining a lower computational cost compared to using high-resolution images directly, there is still a lack of essential components tailored for object detection in aerial images, such as the feature pyramid network.

Feature pyramid networks, which serve as a low-computation alternative to image pyramids, are widely utilized in various detectors and have become essential components in many detection frameworks. The earliest FPN~\cite{lin2017feature} employs a top-down unidirectional path to integrate semantic information into shallow feature maps, effectively enhancing the model's capabilities in multi-scale object detection. Since the unidirectional transmission of information layer by layer inevitably leads to information loss~\cite{chen2020feature}, subsequent feature pyramid networks have progressively shifted towards enabling direct interaction between layers~\cite{tan2020efficientdet, zhang2020feature, yang2023afpn}. However, these structures are typically designed for detecting objects at common scales and often lack the adaptability needed to specific domains, such as small object detection in aerial images. Specifically, these networks typically employ static kernels to process all spatial points across multi-scale feature maps, which is suboptimal for scenes with significant scale variations, as objects of the same category may receive supervision signals across different scales of feature maps. Moreover, applying uniform operations across feature maps of varying scales is not the most effective approach for object detection tasks in aerial images, where numerous small objects are present. 

\begin{figure*}[t]
  \centering
  \includegraphics[width=1\linewidth]{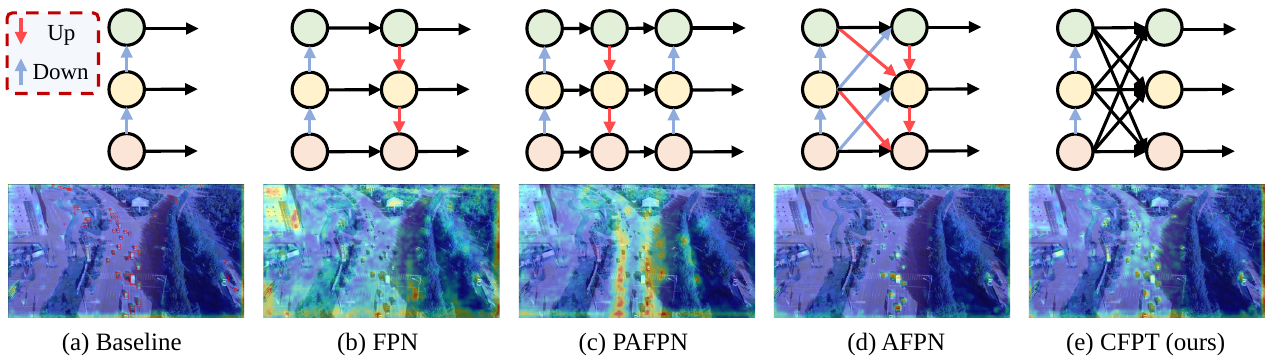}
  \caption{Comparison of the structures and visual feature maps of various feature pyramid networks, including FPN~\cite{lin2017feature}, PAFPN~\cite{liu2018path}, AFPN~\cite{yang2023afpn} and our CFPT. The ``Baseline'' refers to RetinaNet~\cite{lin2017focal} without the feature pyramid network (\ie using vanilla convolutional layers to generate multi-scale feature maps), with red rectangles indicating the ground truths in the current image. 
Our CFPT could effectively focus on multi-scale objects, even those with small scales, while AFPN tends to prioritize larger objects and overlook smaller ones. ``Down" and ``Up" denote downsampling and upsampling operations, respectively. Note that our CFPT does not involve upsampling. Best viewed in color and zoomed in for clarity.}
  \label{fig:structure_comp}
\end{figure*}

\begin{figure}[!t]
  \centering
  \includegraphics[width=0.95\linewidth]{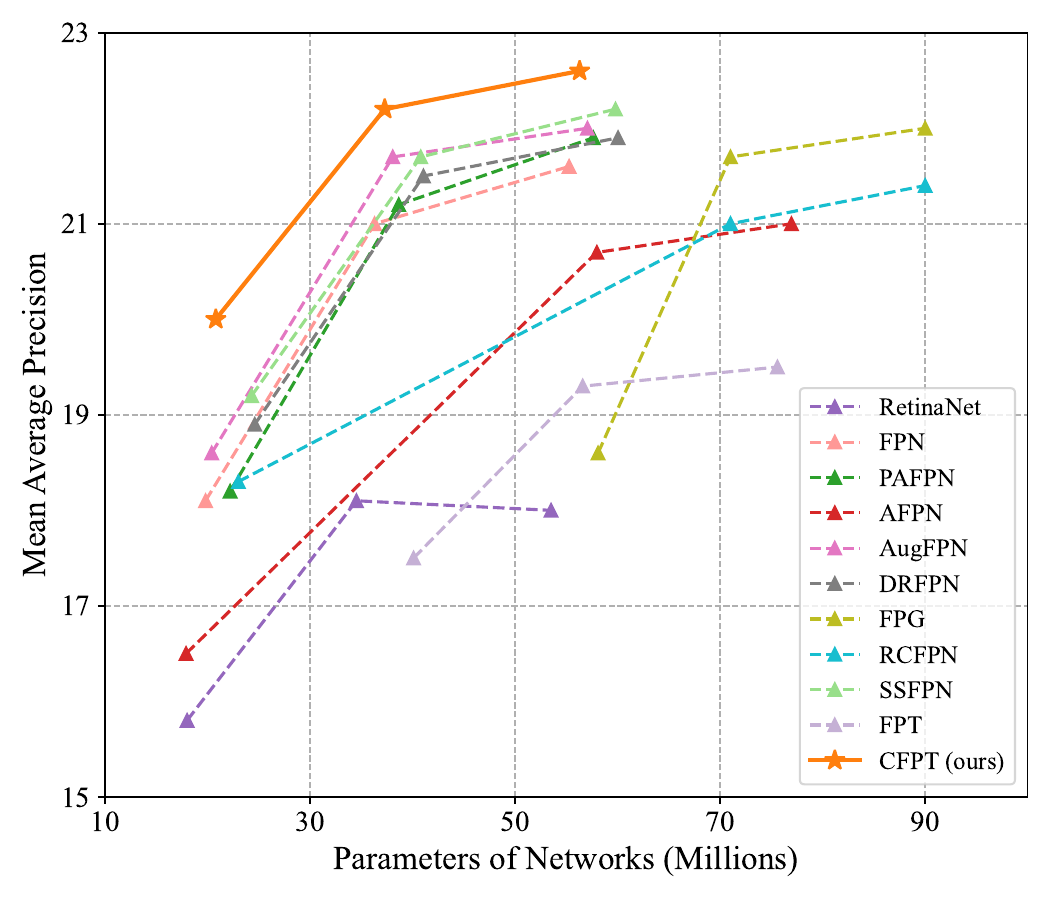}
  \caption{Performance comparison of various state-of-the-art feature pyramid networks on the VisDrone2019-DET dataset. We evaluate their performance by replacing the Neck component in RetinaNet~\cite{lin2017focal}.}
  \label{fig:performance_compare}
\end{figure}

In this paper, we propose a novel feature pyramid network for small object detection in aerial images named Cross-Layer Feature Pyramid Transformer (CFPT). As shown in Fig.~\ref{fig:structure_comp}, CFPT is upsampler-free, which can avoid the additional computation and memory usage associated with upsampling, thereby enhancing computational efficiency. The CFPT consists of two meticulously designed transformer blocks: Cross-Layer Channel-wise Attention (CCA) and Cross-Layer Spatial-wise Attention (CSA). CCA conducts cross-layer feature interactions along the channel dimension, while CSA enables such interactions along the spatial dimension. By integrating these two modules, the model can achieve cross-layer information transfer in a single step, effectively mitigating performance degradation caused by semantic gaps and minimizing the information loss typically associated with layer-by-layer propagation. Furthermore, CFPT effectively integrates global contextual information, which is crucial for detecting small objects, and prioritizes shallow feature maps rich in small objects through adaptive cross-scale neighboring interactions. In addition, to improve the model's positional awareness when capturing global contextual information, we introduce Cross-Layer Consistent Relative Positional Encoding (CCPE), enabling the model to leverage position priors of spatial and channel points across layers, thereby facilitating the derivation of more accurate affinity matrices.

As shown in Fig.~\ref{fig:performance_compare}, our CFPT outperforms other state-of-the-art feature pyramid networks on the VisDrone2019-DET dataset. The proposed CFPT achieves an optimal trade-off between parameters and detection accuracy, demonstrating its effectiveness in detecting small objects in aerial images.

Our contributions can be summarized as follows.
\begin{itemize}
  \item{ 
    We propose CFPT, a novel upsampler-free feature pyramid network for small object detection in aerial images. CFPT can accomplish multi-scale feature interactions in a single step and explicitly provide more attention to shallow feature maps through cross-layer neighborhood interaction groups, achieving lossless information transfer while introducing local inductive bias.
  }
  \item{
    We propose two cross-layer attention blocks with linear computational complexity, namely CCA and CSA, which facilitate cross-layer information interaction in distinct directions (\ie spatial-wise and channel-wise). By integrating both blocks, CFPT can effectively capture the necessary global contextual information for small object detection while maintaining lower computational costs.
  }
  \item{We propose CCPE, a novel positional encoding based on inter-layer mutual receptive fields, designed to enhance the model's awareness of spatial and channel positions during cross-layer interactions.}
  \item{Through extensive experiments on the VisDrone2019-DET, TinyPerson, and xView datasets, we demonstrate the effectiveness of the proposed CFPT for small object detection in aerial images.}
\end{itemize}

\section{Related Work}
\subsection{Small Object Detection in Aerial Images}
Modern object detectors typically reduce the spatial resolution of input images through successive convolutional and pooling layers, aiming to achieve an optimal balance between performance and computational complexity~\cite{ren2015faster, redmon2016you, wang2024yolov10}. Therefore, detecting small objects presents greater challenges than detecting larger ones, as their smaller size amplifies the risk of both information loss and semantic ambiguity during the downsampling process, thereby impacting the performance of object classification and localization.

As an early-stage detection method for small object detection in aerial images, ClusDet~\cite{yang2019clustered} proposes a coarse-to-fine strategy that first identifies dense object clusters and then refines the search within these clusters to enhance the model's ability to detect small objects. Building upon this approach, DMNet~\cite{li2020density} simplifies the training process by employing a density map generation network, which generates density maps to facilitate more efficient cluster prediction. Following a similar detection pipeline, CRENet~\cite{wang2020object} and GLSAN~\cite{deng2020global} further refine the clustering prediction algorithm and optimize the fine-grained prediction process. UFPMP-Det~\cite{huang2022ufpmp} introduces the UFP module and MP-Net to predict sub-regions, which are then reorganized into a new image for more efficient fine-grained detection, thereby improving
both detection accuracy and computational efficiency. CEASC~\cite{du2023adaptive} utilizes sparse convolution to optimize traditional detectors for object detection in aerial images, reducing computational costs while maintaining competitive performance. DTSSNet~\cite{chen2024dtssnet} introduces a manually designed block between the Backbone and Neck to increase the model's sensitivity to multi-scale features and incorporates a training sample selection method specifically for small objects.

The above solutions concentrate on optimizing specific detectors for object detection in aerial images, whereas we propose a new feature pyramid network specifically designed for small object detection in this context.

\subsection{Feature Pyramid Network}
To alleviate the substantial computational costs introduced by image pyramids, feature pyramid networks (FPNs) have been developed as an effective and efficient alternative, improving the performance of various object detectors. FPN~\cite{lin2017feature} utilizes a series of top-down shortcut branches to augment the semantic information lacking in shallow feature maps. Building on the foundation of FPN, PAFPN~\cite{liu2018path} proposes using bottom-up shortcut branches to address the deficiency of detailed information in deep feature maps. Libra-RCNN~\cite{pang2019libra} refines original features by incorporating non-local blocks, enabling the model to obtain more balanced and context-aware interactive features. To bridge the semantic gap in multi-scale feature maps, AugFPN~\cite{guo2020augfpn} introduces a consistent supervision branch and proposes ASF for the dynamic integration of features across multiple scales. FPG~\cite{chen2020feature} represents the feature scale space using a regular grid combined with multi-directional lateral connections among parallel paths, thus enhancing the model's feature representation. AFPN~\cite{yang2023afpn} iteratively refines multi-scale features through the cross-level fusion of deep and shallow feature maps, achieving competitive performance in object detection, particularly in scenarios with common scale distributions.

Unlike previous approaches, we propose CFPT, which leverages global contextual information while explicitly emphasizing the importance of shallow feature maps to enhance the detection performance of small objects in aerial images.

\subsection{Vision Transformer}
As an extension of Transformer~\cite{vaswani2017attention} in the field of computer vision, Vision Transformer (ViT)~\cite{dosovitskiy2020image} has demonstrated significant potential across various visual scenarios~\cite{peng2022spatial, xiao2023enhancing, du2024lda}. Due to the quadratic computational complexity of conventional ViTs regarding image resolution, subsequent research has mainly concentrated on developing lightweight alternatives. Swin Transformer~\cite{liu2021swin} restricts interactions to specific windows and achieves the global receptive field by shifting these windows during the interaction process. 
Local ViTs~\cite{ramachandran2019stand, zhao2020exploring, pan2023slide} incorporate local inductive biases by restricting feature interactions to localized windows, thereby effectively reducing computational complexity and accelerating convergence. Additionally, Axial Attention~\cite{ho2019axial} reduces computational complexity by confining feature interactions to strips along the width and height direction of the image, respectively. 

Inspired by a similar lightweight concept utilized in previous methods, we design two attention blocks with linear complexity (\ie CCA and CSA) to effectively capture the global contextual information across layers in multiple directions (\ie spatial-wise and channel-wise), thereby enhancing the model's ability to detect small objects.

\section{Methodology}
In this section, we provide the technical details of the proposed Cross-layer Feature Pyramid Transformer (CFPT). Section~\ref{subsec:overview} outlines the overall architecture of CFPT. Subsequently, Section~\ref{subsec:cca} and Section~\ref{subsec:csa} introduce the two key components of CFPT: the Cross-layer Channel-wise Attention (CCA) and the Cross-layer Spatial-wise Attention (CSA). Finally, Section~\ref{subsec:ccpe} presents a novel Cross-layer Consistent Relative Positional Encoding (CCPE) designed to enhance the model's cross-layer position-aware capability. 

\begin{figure*}[t]
  \centering
  \includegraphics[width=1\linewidth]{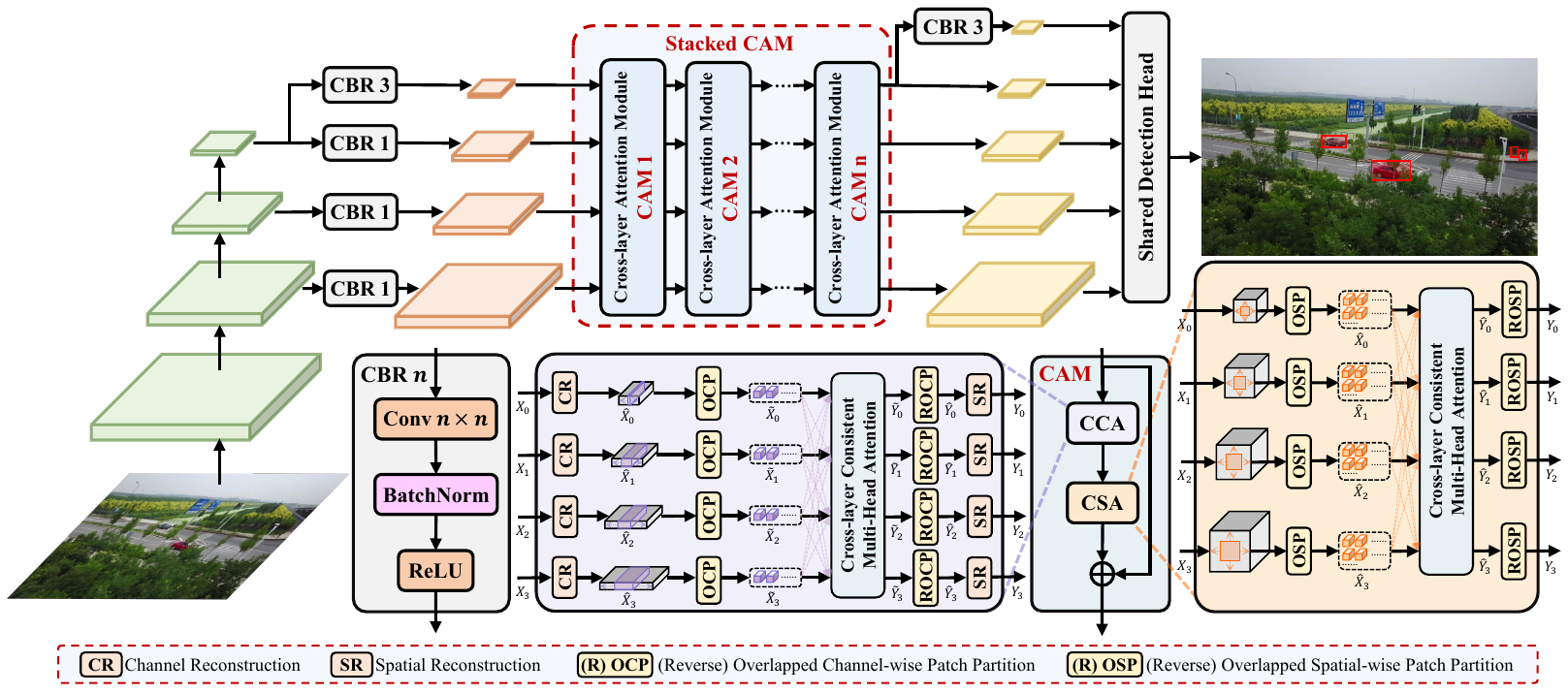}
  \caption{Overall architecture of proposed Cross-layer Feature Pyramid Transformer (CFPT). Given an input image with the shape of $H\times W \times 3$, we apply  Cross-layer Channel-wise Attention (CCA) and Cross-layer Spatial-wise Attention (CSA) multiple times on feature maps downsampled by factors of 8, 16, 32, and 64 to capture cross-layer global contextual information and perform cross-layer adaptive feature correction.}
  \label{fig:overall_archi}
\end{figure*}

\subsection{Overview} \label{subsec:overview}
As illustrated in Fig.~\ref{fig:overall_archi}, CFPT employs multiple parallel CBR blocks to construct inputs for cross-layer feature interaction by leveraging the multi-level feature maps obtained from the feature extraction network (\eg ResNet\cite{he2016deep}), thereby reducing computational complexity and satisfying the architectural requirements of most detectors. Afterward, by leveraging stacked Cross-layer Attention Modules (CAMs), CFPT is able to enhance the model's ability to utilize both global contextual information and cross-layer multi-scale information.

The CAM comprises a sequence of Cross-layer Channel-wise Attention (CCA) and Cross-layer Spatial-wise Attention (CSA). The CCA facilitates local cross-layer interactions along the channel dimension, thereby establishing the global receptive field across the spatial dimension through interactions within each channel-wise token group. Conversely, the CSA enables local cross-layer interactions along the spatial dimension, effectively capturing global contextual information along the channel dimension through interactions within each spatial-wise token group. Furthermore, we incorporate a shortcut connection between the input and output of each CAM to better leverage the benefits of gradient flow.

Assume that the feature map of each scale after the CBR block is represented as $X=\{X_{i}\in \mathcal{R}^{H_{i}\times W_{i}\times C} \}^{L-1}_{i=0}$, where $L$ denotes the number of input layers, and the spatial resolution $H_{i}\times W_{i}$ of each feature map increases with $i$ while the number of channels $C$  remains consistent. The process described above can be expressed as follows: 
\begin{equation}
  Y=\mbox{CAM}(X)=\mbox{CSA}(\mbox{CCA}(X))+X,
\end{equation}
where $Y=\{Y_{i}\in \mathcal{R}^{H_{i}\times W_{i}\times C} \}^{L-1}_{i=0}$ is a set of multi-scale feature maps after cross-layer interaction, maintaining the same shape as the corresponding input feature maps.

It is noteworthy that our CFPT eliminates the complex feature upsampling operations and layer-by-layer information transmission mechanisms, which contribute to increased computational load and memory access delays and are prone to information loss. Instead, we perform a local grouping operation on multi-scale feature maps based on their mutual receptive field sizes, followed by a single-step cross-layer inter-group interaction to facilitate more efficient information exchange across scales. This approach enables features at each scale to acquire information from other layers in a balanced manner, even when the layers are distant, while also facilitating self-correction and benefiting from the inductive bias of locality introduced by local interactions~\cite{pan2023slide}.
\begin{figure}[t]
  \centering
  \includegraphics[width=1\linewidth]{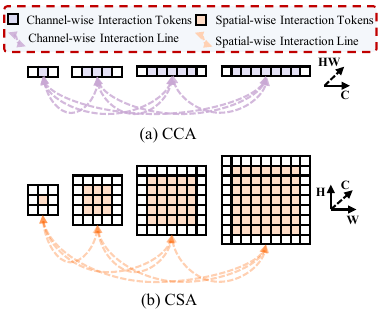}
  \caption{The illustrative diagram of cross-layer neighboring interactions, where blocks of the same color represent tokens at different layers that need to interact. The dashed coordinate system represents the hidden mixed direction of the feature map.}
  \label{fig:cca_csa}
\end{figure}

\subsection{Cross-layer Channel-wise Attention} \label{subsec:cca}
Suppose the input feature maps for the CCA are represented as $X=\{X_{i}\in \mathcal{R}^{H_{i}\times W_{i}\times C} \}^{L-1}_{i=0}$. As shown in Fig.~\ref{fig:cca_csa}(a), the CCA performs multi-scale neighboring interactions across layers along the channel dimension, thereby providing global contextual information along the spatial dimension for each channel-wise token. 
To construct the interactive inputs, Channel Reconstruction (CR) is initially applied to the feature map of each scale to ensure consistent spatial resolution, resulting in $\hat{X}=\{\hat{X}_{i}\in \mathcal{R}^{H_{0}\times W_{0} \times \frac{H_{i}W_{i}}{H_{0}W_{0}} \times C} \}^{L-1}_{i=0}$. CR functions similarly to Focus in YOLOv5 but eliminates the need for additional feature mapping operations. Instead, CR stacks feature values from the spatial dimension into the channel dimension, achieving consistent spatial resolution while maintaining efficiency. The above process can be described as
\begin{equation}
  \hat{X}=\mbox{CR}(X).
\end{equation}

Next, we apply the Overlapped Channel-wise Patch Partition (OCP) to generate channel-wise token groups, which can be viewed as the Patch Embedding~\cite{dosovitskiy2020image} that extracts overlapping regions from local areas along the channel dimension, with patch sizes varying across feature maps at different scales. Specifically, based on the shape of multi-scale features, the channel sizes of adjacent feature maps in $\hat{X}$ differ by a factor of $4$ (\ie $\frac{H_{i}W_{i}}{H_{0}W_{0}}=4^{i}$). To construct overlapping neighboring interaction groups, we introduce an expansion factor $\alpha$ and perform OCP on $\hat{X}$, resulting in $\tilde{X}=\{\tilde{X}_{i} \in\mathcal{R}^{C\times (4^{i}+\alpha)\times H_{0}W_{0}}\}^{L-1}_{i=0}$, which can be described as
\begin{equation}
  \tilde{X}=\mbox{OCP}(\hat{X}, \alpha).
\end{equation}

Taking the feature map of the $i$-th layer as an example, after obtaining $\tilde{X}_{i}$, we employ the cross-layer consistent multi-head attention to capture the global dependencies along the spatial dimension, thereby obtaining the interaction result $\tilde{Y}_{i} \in\mathcal{R}^{C\times (4^{i}+\alpha)\times H_{0}W_{0}}$. The above process can be expressed as
\begin{align}
  (Q_{i}, K_{i}, V_{i})&=(W^{q}_{i}\tilde{X}_{i}, W^{k}_{i}\tilde{X}_{i}, W^{v}_{i}\tilde{X}_{i}), \\
  \tilde{Y}_{i}&=\mbox{Softmax}(\frac{Q_{i}K^{T}}{\sqrt{d_{k}}})V \label{equ:qkv},
\end{align}
where $W^{q}_{i}, W^{k}_{i}, W^{v}_{i} \in \mathcal{R}^{C\times C}$ are the linear projection matrices. $K= [K_{0}, K_{1}, \ldots, K_{L-1}] \in \mathcal{R}^{C\times \sum_{i=0}^{L-1}(4^{i}+\alpha)\times H_{0}W_{0}}$ and $V = [V_{0}, V_{1}, \ldots, V_{L-1}]\in \mathcal{R}^{C\times \sum_{i=0}^{L-1}(4^{i}+\alpha)\times H_{0}W_{0}}$ represent the concatenated keys and values, respectively, where $[\cdot]$ represents concatenation operation. Note that for simplicity, we consider the case where the number of heads is $1$. In practice, we employ the multi-head mechanism to capture global dependencies for each channel-wise token.

After obtaining the interaction results $\tilde{Y} = \{\tilde{Y}_{i}\}^{L-1}_{i=0}$ for feature maps at each scale, we apply the Reverse Overlapped Channel-wise Patch Partition (ROCP) to restore the impact of OCP and recover the original spatial resolution. This results in $\hat{Y}=\{\hat{Y}_{i} \in \mathcal{R}^{H_{0}\times W_{0} \times \frac{H_{i}W_{i}}{H_{0}W_{0}} \times C}\}^{L-1}_{i=0}$. As the reverse operation of OCP, ROCP restores the spatial resolution using the same kernel size and stride as those applied in OCP.

Finally, we apply Spatial Reconstruction (SR), which reverses the operation of CR by rearranging feature values from the channel dimension back into the spatial dimension, producing the output $Y=\{Y_{i}\in \mathcal{R}^{H_{i}\times W_{i}\times C} \}^{L-1}_{i=0}$, ensuring that the shape of each feature map aligns with the corresponding input $X_{i}$. This process can be formulated as
\begin{equation}
  Y=\mbox{SR}(\hat{Y})=\mbox{SR}(\mbox{ROCP}(\tilde{Y}, \alpha)).
\end{equation}

\subsection{Cross-layer Spatial-wise Attention} \label{subsec:csa}
Similarly, let the set of input feature maps for the CSA be denoted as $X = \{X_{i} \in \mathcal{R}^{H_{i} \times W_{i} \times C}\}_{i=0}^{L-1}$. As illustrated in Fig.~\ref{fig:cca_csa}(b), CSA performs multi-scale neighboring interactions across layers along the spatial dimension, thus providing global contextual information along the channel dimension for each spatial-wise token. 

Since the channel sizes of the input feature maps are matched after the CBR block(\eg 256), there is no need for additional alignment operations such as CR and SR, which are required in CCA. Therefore, we can directly perform Overlapped Spatial-wise Patch Partition (OSP) to obtain spatial-wise token groups. This operation can be interpreted as applying sliding crops on feature maps of different scales using rectangular boxes of varying sizes. Assuming the expansion factor for OSP is $\beta$, this operation produces $\hat{X} = \{\hat{X}_{i} \in \mathcal{R}^{H_{0}\times W_{0} \times (2^{i}+\beta)^{2} \times C}\}_{i=0}^{L-1}$, which can be expressed as
\begin{equation}
  \hat{X}=\mbox{OSP}(X, \beta).
\end{equation}

Then, we can perform local interaction within cross-layer spatial-wise token groups and use the cross-layer consistent multi-head attention to capture global dependencies along the channel dimension, thereby obtaining $\hat{Y}=\{\hat{Y}_{i} \in\mathcal{R}^{H_{0}\times W_{0} \times (2^{i}+\beta)^{2} \times C}\}_{i=0}^{L-1}$. Specifically, for the feature map at the $i$-th layer, this process can be expressed as follows:
\begin{align}
  (Q_{i}, K_{i}, V_{i})&=(W^{q}_{i}\hat{X}_{i}, W^{k}_{i}\hat{X}_{i}, W^{v}_{i}\hat{X}_{i}), \\
  \hat{Y}_{i}&=\mbox{Softmax}(\frac{Q_{i}K^{T}}{\sqrt{d_{k}}})V ,\label{equ:qkv2}
\end{align}
where $W^{q}_{i}, W^{k}_{i}, W^{v}_{i} \in \mathcal{R}^{C\times C}$ represent the linear projection matrices. $K= [K_{0}, K_{1}, ..., K_{L-1}]\in \mathcal{R}^{H_{0} \times W_{0} \times \sum_{i=0}^{L-1}(2^{i}+\beta)^{2}\times C}$ and $V = [V_{0}, V_{1}, ..., V_{L-1}]\in \mathcal{R}^{H_{0} \times W_{0} \times \sum_{i=0}^{L-1}(2^{i}+\beta)^{2}\times C}$ denote the concatenated keys and values across all layers, respectively.

Next, we employ Reverse Overlapped Spatial Patch Partition (ROSP) to reverse the effect of OSP, thereby obtaining the interaction results $Y = \{Y_{i} \in \mathcal{R}^{H_{i} \times W_{i} \times C}\}_{i=0}^{L-1}$:
\begin{equation}
  Y=\mbox{ROSP}(\hat{Y}, \beta).
\end{equation}

\subsection{Cross-layer Consistent Relative Positional Encoding} \label{subsec:ccpe}
Since each token-pair within its cross-layer token group maintains specific positional relationships during the interaction process. However, the vanilla multi-head attention treats all interaction tokens uniformly, resulting in suboptimal results for position-sensitive tasks such as object detection. Therefore, we introduce Cross-layer Consistent Relative Positional Encoding (CCPE) to enhance the cross-layer location awareness of CFPT during the interaction process.
\begin{figure}[t]
  \centering
  \includegraphics[width=1\linewidth]{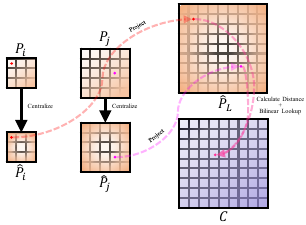}
  \caption{The flowchart of Cross-layer Consistent Relative Positional Encoding, where ``Bilinear Lookup" indicates table lookup based on non-integer indices using bilinear interpolation.}
  \label{fig:ccpe}
\end{figure}

The core idea of CCPE is based on the aligned mutual receptive fields across multiple scales, driven by the theory of receptive field expansion in convolution~\cite{simonyan2014very}. Taking CSA as an example, the set of attention maps between each spatial-wise token group can be expressed as $A=\{A_{i,j}\in \mathcal{R}^{N \times H_{0} \times W_{0} \times (2^{i}+\beta)^{2} \times (2^{j}+\beta)^{2}}\}_{i,j=0}^{L-1}$, where $N$ is the number of heads and $A_{i,j} = Q_{i}K_{j}^{T} / \sqrt{d_k}$, as defined in Equation~\ref{equ:qkv2}. For simplicity, we ignore $H_{0}$ and $W_{0}$ and define $s_i^h=s_i^w = 2^{i}+\beta$ and $s_j^h=s_j^w=2^{j}+\beta$, where $(s_i^h, s_i^w)$ and $(s_j^h, s_j^w)$ represent the height and width of the spatial-wise token groups in the $i$-th and $j$-th layers. Therefore, the set of attention maps can be re-expressed as $A=\{A_{i,j}\in \mathcal{R}^{N \times s_i^hs_i^w \times s_j^hs_j^w}\}_{i,j=0}^{L-1}$.

The process of CCPE is shown Fig.~\ref{fig:ccpe}. We define a learnable codebook $C\in \mathcal{R}^{N \times (2s_{L-1}^{h}-1) \times (2s_{L-1}^{w} - 1)}$ and extract the relative position information between any two tokens from the codebook by calculating their cross-layer consistent relative position index. For simplicity, consider the interaction of the spatial-wise token groups from the $i$-th and $j$-th layers, where $P_i \in \mathcal{R}^{s^{h}_{i} \times s^{w}_{i} \times 2}$ and $P_j \in \mathcal{R}^{s^{h}_{j} \times s^{w}_{j} \times 2}$ represent their respective absolute coordinate matrices, which can be expressed as
\begin{equation}
  \left\{
  \begin{aligned}
    P_i &= \left\{(p_i^w, p_i^h) \mid p_i^w \in [0, s_i^w - 1], p_i^h \in [0, s_i^h - 1] \right\} \\
    P_j &= \left\{(p_j^w, p_j^h) \mid p_j^w \in [0, s_j^w - 1], p_j^h \in [0, s_j^h - 1] \right\}. \\
  \end{aligned}
  \right.
\end{equation}

To obtain the relative position information of $P_{j}$ relative to $P_{i}$, we first centralize their coordinates using their respective spatial-wise token group sizes to obtain $\hat{P}_{i}$ and $\hat{P}_{j}$: 
\begin{equation}
  \left\{
\begin{aligned}
  \hat{P}_i&=\left\{(\hat{p}_i^w, \hat{p}_i^h) \mid \hat{p}_i^w=p_{i}^{w} - \left\lfloor \frac{s^w_i}{2}\right\rfloor, \hat{p}_i^h=p_{i}^{h} - \left\lfloor \frac{s^h_i}{2}\right\rfloor\right\} \\
  \hat{P}_j&=\left\{(\hat{p}_j^w, \hat{p}_j^h) \mid \hat{p}_j^w=p_{j}^{w} - \left\lfloor \frac{s^w_j}{2}\right\rfloor, \hat{p}_j^h=p_{j}^{h} - \left\lfloor \frac{s^h_j}{2}\right\rfloor\right\}. \\
\end{aligned}
  \right.
\end{equation}

Then we can derive $\tilde{P}_i$ and $\tilde{P}_j$ by projecting their coordinates to the largest spatial-wise token group:
\begin{equation}\label{equ:proj}
  \left\{
\begin{aligned}
  \tilde{P}_i&=\left\{(\tilde{p}_i^w, \tilde{p}_i^h) \mid \tilde{p}_i^w=\frac{\hat{p}_{i}^{w}s_{L-1}^{w}}{s_i^w}, \tilde{p}_i^h=\frac{\hat{p}_{i}^{h}s_{L-1}^{h}}{s_i^h} \right\} \\
  \tilde{P}_j&=\left\{(\tilde{p}_j^w, \tilde{p}_j^h) \mid \tilde{p}_j^w=\frac{\hat{p}_{j}^{w}s_{L-1}^{w}}{s_j^w}, \tilde{p}_j^h=\frac{\hat{p}_{j}^{h}s_{L-1}^{h}}{s_j^h} \right\}. \\
\end{aligned}
  \right.
\end{equation}

Subsequently, the relative distance between $\tilde{P}_i$ and $\tilde{P}_j$ can be calculated and converted the relative distance into the index of the codebook using the following equations:
\begin{equation}
  \left\{
\begin{aligned}
  & D_{i,j} = \left\{(d_{i,j}^w, d_{i,j}^h) \mid d_{i,j}^w=\tilde{p}_i^w - \tilde{p}_j^w, d_{i,j}^h=\tilde{p}_i^h - \tilde{p}_j^h \right\} \\
  & I_{i,j} = \left\{(I_{i,j}^w, I_{i,j}^h) \mid I_{i,j}^w=d_{i,j}^{w} + s_{L}^{w} - 1, I_{i,j}^h=d_{i,j}^{h} + s_{L}^{h} - 1 \right\}.
\end{aligned}
\right.
\end{equation}

Finally, the positional embedding matrix is retrieved from the codebook $C$ using the index $I_{i,j}$ and superimposed onto the original attention map $A_{i,j}$, resulting in the output $\hat{A}_{i,j}$ with enriched positional information. This positional information can be incorporated into the attention mechanism by modifying Equation~\ref{equ:qkv2} as follows:
\begin{align}
  \hat{Y}_{i}&=\mbox{Softmax}(\hat{A}_{i,j})V, \\
  &=\mbox{Softmax}(A_{i,j}+f(C,I_{i,j}))V, 
\end{align}
where $f(a, b)$ is the bilinear interpolation function used to ensure the differentiability while extracting relative positional information from the codebook using non-integer coordinates derived from Equation~\ref{equ:proj}, with $a$ representing the codebook and $b$ denoting the non-integer coordinates matrix. Note that the process of introducing CCPE into CCA is the same as that for CSA, and will not be discussed in detail here.

\begin{figure}[t]
  \centering
  \includegraphics[width=1\linewidth]{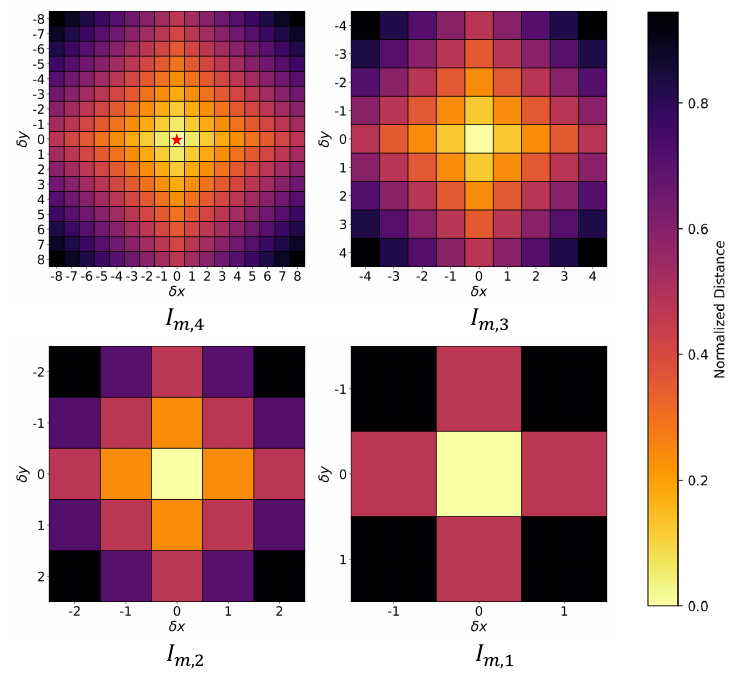}
  \caption{Visualization of cross-layer bilinear lookup index matrices. \textcolor{red}{\ding{72}} indicates the query point, located at the center of the fifth layer (denoted as point $m$). Each matrix represents the bilinear lookup indices for the spatial interaction group corresponding to the query point $m$ in a specific layer $n$, where $n \in [0, L-1]$. For simplicity, we suppose $L=5$ and exclude $n=0$ from the visualization. The colors represent the Euclidean distance between position indices, where similar colors correspond to smaller relative distances.}
  \label{fig:ccpe_vis}
\end{figure}

\noindent\textbf{Visualization.} To provide a clearer understanding of CCPE, we use the center of the fifth layer (denoted as position $m$) as an example to visualize the generated cross-layer position index matrices $I_{m}=\{I_{m,n}\in \mathcal{R}^{s^{h}_{n} \times s^{w}_{n}}\}_{n=0}^{L-1}$. These matrices represent the bilinear lookup indices of all points within the spatial interaction group associated with the $n$-th layer. As shown in Fig.~\ref{fig:ccpe_vis}, CCPE is performed based on the consistent receptive field across layers, ensuring that spatially close points across layers have position indices that are also close.

\subsection{Complexity Analysis} \label{subsec:complexity}
In this section, we will analyze the computational complexity of both CCA and CSA. Note that the sizes of the channel-wise and spatial-wise token groups of CCA and CSA remain constant during both training and testing phases, ensuring their computational complexity scales linearly with the spatial resolution of the input feature maps.
\subsubsection{Cross-layer Channel-wise Attention}
Consider a set of multi-scale input feature maps denoted as $X=\{X_{i}\in \mathcal{R}^{H_{i}\times W_{i}\times C} \}^{L-1}_{i=0}$, with $\alpha$ denoting the expansion factor utilized in CCA. The overall computational complexity of CCA includes $\mathcal{O}(\sum_{i = 0}^{L-1}(4H_{i}W_{i}C^{2}))$ for linear projections, $\mathcal{O}(\sum_{i = 0}^{L-1}(2C(4^{i}+\alpha)^{2}H_{0}W_{0}))$ for attention interactions, and $\mathcal{O}(\sum_{i = 0}^{L-1}(8H_{i}W_{i}C^{2}))$ for Feed-Forward Networks (FFNs).

\subsubsection{Cross-layer Spatial-wise Attention}
Suppose the set of multi-scale input feature maps is $X=\{X_{i}\in \mathcal{R}^{H_{i}\times W_{i}\times C} \}^{L-1}_{i=0}$, with $\beta$ denoting the expansion factor applied in CSA. The overall computational complexity of CSA includes $\mathcal{O}(\sum_{i = 0}^{L-1}(4H_{i}W_{i}C^{2}))$ for linear projections, $\mathcal{O}(\sum_{i = 0}^{L-1}(2C(2^{i}+\beta)^{4}H_{0}W_{0}))$ for attention interactions, and $\mathcal{O}(\sum_{i = 0}^{L-1}(8H_{i}W_{i}C^{2}))$ for FFNs.

\begin{table*}[t!]
  \centering
    \caption{Performance comparison of various state-of-the-art feature pyramid networks on the VisDrone2019-DET dataset. RetinaNet is employed as the baseline model to evaluate the performance of various feature pyramid networks by replacing the Neck component. The best performance is highlighted in bold.}
  \renewcommand{\arraystretch}{1.1}
    \label{tab:vari_fpn}
    \resizebox{\linewidth}{!}{
    \begin{tabular}{l|c|c|ccc|ccc|cc}
      \toprule
      Method& Backbone &Schedule & AP(\%) & AP$_{0.5}$(\%) & AP$_{0.75}$(\%) & AP-small(\%) & AP-medium(\%) & AP-large(\%) & Params(M) & FLOPs(G) \\
      \midrule
      RetinaNet~\cite{lin2017focal} & ResNet-18 & 1$\times$& 15.8 & 28.4 & 15.6 & 7.5 & 24.4 & 33.2 & 18.0 & 151.1 \\
      FPN~\cite{lin2017feature} & ResNet-18 & 1$\times$&18.1 & 32.7 & 17.8 & 9.2 & 28.7 & 33.6 & 19.8 & 164.0 \\
      PAFPN~\cite{liu2018path} & ResNet-18 & 1$\times$&18.2 & 32.5 & 18.2 & 8.9 & 28.9 & 36.3 & 22.2 & 170.1 \\
      AugFPN~\cite{guo2020augfpn} & ResNet-18 & 1$\times$&18.6 & 33.2 & 18.4 & 9.0 & 29.5 & 37.3 & 20.4 & 164.2 \\
      DRFPN~\cite{ma2020dual} & ResNet-18 & 1$\times$&18.9 & 33.4 & 18.8 & 9.1 & 30.2 & \textbf{38.5} & 24.6 & 176.0 \\
      FPG~\cite{chen2020feature} & ResNet-18 & 1$\times$&18.6 & 33.2 & 18.4 & 9.5 & 29.6 & 36.0 & 58.1 & 290.5 \\
      FPT~\cite{zhang2020feature} & ResNet-18 & 1$\times$& 17.5 & 30.7 & 17.5 & 8.3 & 28.0 & 37.9 & 40.1 & 275.2 \\
      RCFPN~\cite{zong2021rcnet} & ResNet-18 & 1$\times$&18.3 & 32.4 & 18.1 & 8.6 & 29.3 & 36.3 & 23.0 & 157.5 \\
      SSFPN~\cite{hong2021sspnet} & ResNet-18 & 1$\times$&19.2 & 33.7 & 19.1 & 10.0 & 31.2 & 35.8 & 24.3 & 221.4 \\
      AFPN~\cite{yang2023afpn} & ResNet-18 & 1$\times$&16.5 & 30.0 & 16.5 & 8.2 & 26.0 & 32.3 & 17.9 & 153.2 \\
      CFPT (ours) & ResNet-18 & 1$\times$&\textbf{20.0} & \textbf{35.3} & \textbf{20.0} & \textbf{10.1} & \textbf{31.7} & 37.2 & 20.8 & 165.9 \\
      \midrule
      RetinaNet~\cite{lin2017focal} & ResNet-50 & 1$\times$& 18.1 & 31.1 & 18.3 & 8.8 & 28.5 & 38.0 & 34.5 & 203.7 \\
      FPN~\cite{lin2017feature} & ResNet-50 & 1$\times$& 21.0 & 36.4 & 21.4 & 10.9 & 34.3 & 40.1 & 36.3 & 216.6 \\
      PAFPN~\cite{liu2018path} & ResNet-50 & 1$\times$& 21.2 & 36.5 & 21.6 & 10.9 & 34.6 & 41.1 & 38.7 & 222.7 \\
      AugFPN~\cite{guo2020augfpn} & ResNet-50 & 1$\times$& 21.7 & 37.1 & 22.2 & 11.1 & \textbf{35.4} & 40.4 & 38.1 & 216.8\\
      DRFPN~\cite{ma2020dual} & ResNet-50 & 1$\times$& 21.5 & 36.7 & 22.0 & 11.0 & 35.3 & 39.5 & 41.1 & 228.5 \\
      FPG~\cite{chen2020feature} & ResNet-50 & 1$\times$& 21.7 & 37.3 & 22.2 & 11.5 & 35.2 & 38.7 & 71.0 & 346.1 \\
      FPT~\cite{zhang2020feature} & ResNet-50 & 1$\times$& 19.3 & 33.3 & 19.2 & 9.4 & 30.0 & 38.9 & 56.6 & 331.8\\
      RCFPN~\cite{zong2021rcnet} & ResNet-50 & 1$\times$& 21.0 & 36.0 & 21.3 & 10.5 & 34.8 & 38.1 & 36.0 & 209.2 \\
      SSFPN~\cite{hong2021sspnet} & ResNet-50 & 1$\times$& 21.7 & 37.3 & 22.2 & 11.5 & 35.3 & 39.8 & 40.8 & 274.0 \\
      AFPN~\cite{yang2023afpn} & ResNet-50 & 1$\times$& 20.7 & 36.0 & 21.2 & 10.7 & 33.4 & 36.9 & 58.0 & 250.0 \\
      CFPT (ours) & ResNet-50 & 1$\times$& \textbf{22.2} & \textbf{38.0} & \textbf{22.4} & \textbf{11.9} & 35.2 & \textbf{41.7} & 37.3 & 218.5 \\
      \midrule
      RetinaNet~\cite{lin2017focal} & ResNet-101 & 1$\times$& 18.0 & 31.0 & 18.3 & 8.8 & 28.5 & 38.0 & 53.5 & 282.8\\
      FPN~\cite{lin2017feature} & ResNet-101 & 1$\times$& 21.6 & 37.3 & 21.8 & 11.2 & 34.9 & 41.9 & 55.3 & 295.7 \\
      PAFPN~\cite{liu2018path} & ResNet-101 & 1$\times$& 21.9 & 37.4 & 22.2 & 11.6 & 35.4 & 42.5 & 57.6 & 301.8 \\
      AugFPN~\cite{guo2020augfpn} & ResNet-101 & 1$\times$& 22.0 & 37.8 & 22.4 & 11.3 & 36.0 & 43.2 & 57.1 & 296.0 \\
      DRFPN~\cite{ma2020dual} & ResNet-101 & 1$\times$& 22.0 & 37.8 & 22.4 & 11.5 & 36.0 & 41.1 & 60.1 & 307.7 \\
      FPG~\cite{chen2020feature} & ResNet-101 & 1$\times$& 22.0 & 37.9 & 22.4 & 11.5 & 35.7 & 42.0 & 90.0 & 431.3 \\
      FPT~\cite{zhang2020feature} & ResNet-101 & 1$\times$& 19.5 & 33.5 & 19.9 & 9.4 & 30.5 & 39.8 & 75.6 & 417.0\\
      RCFPN~\cite{zong2021rcnet} & ResNet-101 & 1$\times$& 21.4 & 36.8 & 21.7 & 11.1 & 35.2 & 40.0 & 55.0 & 288.3 \\
      SSFPN~\cite{hong2021sspnet} & ResNet-101 & 1$\times$& 22.2 & 38.3 & 22.6 & 11.9 & 35.8 & 43.3 & 59.8 & 353.1 \\
      AFPN~\cite{yang2023afpn} & ResNet-101 & 1$\times$& 21.0 & 36.7 & 21.6 & 11.2 & 33.7 & 36.7 & 77.0 & 329.1 \\
      CFPT (ours) & ResNet-101 & 1$\times$& \textbf{22.6} & \textbf{38.4} & \textbf{23.1} & \textbf{12.1} & \textbf{36.2} & \textbf{43.8} & 56.3 & 297.6\\
      \midrule
      RetinaNet~\cite{lin2017focal} & ResNet-18 & 3$\times$& 19.6 & 34.2 & 19.8 & 9.9 & 30.5 & 36.9 & 18.0 & 151.1 \\
      FPN~\cite{lin2017feature} & ResNet-18 & 3$\times$& 21.4 & 37.3 & 21.6 & 11.4 & 33.6 & 40.1 & 19.8 & 164.0 \\
      PAFPN~\cite{liu2018path} & ResNet-18 & 3$\times$ & 21.3 & 37.1 & 21.5 & 11.1 & 33.8 & 40.5 & 22.2 & 170.1 \\
      AugFPN~\cite{guo2020augfpn} & ResNet-18 & 3$\times$ & 21.9 & 38.0 & 21.9 & 11.3 & 34.8 & 41.7 & 20.4 & 164.2 \\
      DRFPN~\cite{ma2020dual} & ResNet-18 & 3$\times$ & 22.4 & 38.2 & 22.7 & 11.9 & 35.5 & 41.2 & 24.6 & 176.0 \\
      FPG~\cite{chen2020feature} & ResNet-18 & 3$\times$ & 21.9 & 38.5 & 22.4 & \textbf{12.4} & 34.0 & 40.3 & 58.1 & 290.5 \\
      FPT~\cite{zhang2020feature} & ResNet-18 & 3$\times$ & 21.1 & 36.7 & 21.5 & 11.1 & 33.3 & \textbf{42.2} & 40.1 & 275.2 \\
      RCFPN~\cite{zong2021rcnet} & ResNet-18 & 3$\times$ & 22.4 & 38.5 & 22.5 & 11.8 & 35.7 & 42.1 & 23.0 & 157.5 \\
      SSFPN~\cite{hong2021sspnet} & ResNet-18 & 3$\times$ & 22.5 & 38.7 & 23.0 & 12.1 & 35.5 & 38.4 & 24.3 & 221.4 \\
      AFPN~\cite{yang2023afpn} & ResNet-18 & 3$\times$ & 19.1 & 34.3 & 19.2 & 9.8 & 29.7 & 34.0 & 17.9 & 153.2 \\
      CFPT (ours) & ResNet-18 & 3$\times$ & \textbf{22.9} & \textbf{39.4} & \textbf{23.2} & \textbf{12.4} & \textbf{36.2} & 40.9 & 20.8 & 165.9 \\
       \midrule
      RetinaNet~\cite{lin2017focal} & ResNet-50 & 3$\times$& 22.5 & 37.9 & 22.9 & 11.9 & 34.8 & 40.7 & 34.5 & 203.7 \\
      FPN~\cite{lin2017feature} & ResNet-50 & 3$\times$ & 23.5 & 40.0 & 24.1 & 12.5 & 37.2 & 42.8 & 36.3 & 216.6\\
      PAFPN~\cite{liu2018path} & ResNet-50 & 3$\times$ & 23.7 & 40.0 & 24.4 & 13.0 & 37.3 & 43.0 & 38.7 & 222.7 \\
      AugFPN~\cite{guo2020augfpn} & ResNet-50 & 3$\times$ & 24.1 & 40.8 & 24.3 & 13.3 & 37.7 & 43.8 & 38.1 & 216.8 \\
      DRFPN~\cite{ma2020dual} & ResNet-50 & 3$\times$ & 24.0 & 40.7 & 24.4 & 13.2 & 37.8 & 43.8 & 41.1 & 228.5 \\
      FPG~\cite{chen2020feature} & ResNet-50 & 3$\times$ & 24.2 & 41.1 & 25.1 & 13.8 & 37.5 & 41.7 & 71.0 & 346.1 \\
      FPT~\cite{zhang2020feature} & ResNet-50 & 3$\times$ & 23.1 & 39.2 & 23.4 & 12.6 & 35.5 & 43.1 & 56.6 & 331.8  \\
      RCFPN~\cite{zong2021rcnet} & ResNet-50 & 3$\times$ & 24.2 & 40.8 & 24.9 & 13.2 & 38.2 & 42.2 & 36.0 & 209.2 \\
      SSFPN~\cite{hong2021sspnet} & ResNet-50 & 3$\times$ & 24.3 & 40.4 & 25.0 & 13.4 & 37.7 & 42.2 & 40.8 & 274.0 \\
      AFPN~\cite{yang2023afpn} & ResNet-50 & 3$\times$ & 24.2 & 41.0 & 24.4 & 13.3 & 37.9 & 44.4 & 58.0 & 250.0 \\
      CFPT (ours) & ResNet-50 & 3$\times$ & \textbf{25.0} & \textbf{42.0} & \textbf{25.5} & \textbf{13.9} & \textbf{38.6} & \textbf{44.9} & 37.3 & 218.5 \\
      \bottomrule
    \end{tabular}}
  \end{table*}

  \begin{table*}[t]
    \centering
      \caption{Performance comparison with other state-of-the-art detectors on the VisDrone2019-DET dataset. The best performance is highlighted in bold. $^{\dag}$ indicates performance cited from other work. }
    \renewcommand{\arraystretch}{1.1}
      \label{tab:vari_det}
      \resizebox{\linewidth}{!}{
      \begin{tabular}{l|c|ccc|ccc|cc}
        \toprule
        Method & Backbone & AP(\%) & AP$_{0.5}$(\%) & AP$_{0.75}$(\%) & AP-small(\%) & AP-medium(\%) & AP-large(\%) & Params(M) & FLOPs(G) \\
        \midrule
        \textit{Two-stage object detectors:} \\
        Faster RCNN~\cite{ren2015faster} & ResNet-18 & 22.0 & 38.4 & 22.5 & 13.7 & 32.0 & 37.3 & 28.2 & 159.5\\
        Faster RCNN~\cite{ren2015faster} & ResNet-50 & 25.0 & 42.4 & 25.8 & 15.6 & 36.5 & 43.0 & 41.2 & 214.4\\
        Faster RCNN~\cite{ren2015faster} & ResNet-101 & 25.2 & 42.7 & 26.8 & 15.8 & 36.9 & 41.9 & 60.2 & 293.5\\
        Cascade RCNN~\cite{cai2018cascade} & ResNet-18 & 23.3 & 39.2 & 23.9 & 14.3 & 34.1 & 44.3 & 56.0 & 187.3\\
        Cascade RCNN~\cite{cai2018cascade} & ResNet-50 & 25.9 & 42.6 & 27.4 & 15.8 & 38.1 & 45.2 & 69.0 & 242.2\\
        Cascade RCNN~\cite{cai2018cascade} & ResNet-101 & 26.0 & 42.8 & 27.3 & 16.1 & 38.2 & 50.4 & 88.0 & 321.3\\
        Libra RCNN~\cite{pang2019libra} & ResNet-18 & 21.9 & 36.9 & 22.8 & 14.1 & 32.3 & 37.9 & 28.4 & 160.6\\
        Libra RCNN~\cite{pang2019libra} & ResNet-50 & 25.2 & 41.7 & 26.5 & 15.9 & 36.6 & 42.2 & 41.4 & 215.5\\
        Libra RCNN~\cite{pang2019libra} & ResNet-101 & 25.3 & 42.4 & 26.4 & 16.0 & 37.2 & 41.5 & 60.4 & 294.6\\
        PISA~\cite{cao2020prime} & ResNet-18 & 23.7 & 40.1 & 24.7 & 15.1 & 34.0 & 39.1 & 28.2 & 159.5 \\
        PISA~\cite{cao2020prime} & ResNet-50 & 26.6 & 44,2 & 27.6 & 16.7 & 38.4 & 43.5 & 41.2 & 214.4 \\
        PISA~\cite{cao2020prime} & ResNet-101 & 26.4 & 44.2 & 27.9 & 17.2 & 37.6 & 45.4 & 60.2 & 293.5 \\
        Dynamic RCNN~\cite{zhang2020dynamic} & ResNet-18  & 16.9 & 28.9 & 17.3 & 10.2 & 25.1 & 28.5 & 28.2 & 159.5  \\
        Dynamic RCNN~\cite{zhang2020dynamic} & ResNet-50  & 19.5 & 32.2 & 20.4 & 12.5 & 28.3 & 37.2 & 41.2 & 214.4 \\
        Dynamic RCNN~\cite{zhang2020dynamic} & ResNet-101  & 19.2 & 31.6 & 20.0 & 12.4 & 28.0 & 38.0 & 60.2 & 293.5 \\
        \midrule
        \textit{One-stage object detectors:} \\
        RetinaNet~\cite{lin2017focal} & ResNet-18 & 19.3 & 34.3 & 19.2 & 9.7 & 30.9 & 37.9 & 19.8 & 164.0\\
        RetinaNet~\cite{lin2017focal} & ResNet-50 & 22.5 & 38.7 & 23.0 & 11.5 & 36.2 & 42.3 & 36.3 & 216.6\\
        RetinaNet~\cite{lin2017focal} & ResNet-101 & 23.0 & 39.2 & 23.5 & 11.7 & 37.2 & 43.9 & 55.3 & 295.7 \\
        FSAF~\cite{zhu2019feature} & ResNet-18 & 19.3 & 35.1 & 18.3 & 11.7 & 27.5 & 34.2 & 19.6 & 158.3 \\
        FSAF~\cite{zhu2019feature} & ResNet-50 & 22.9 & 40.2 & 22.6 & 14.1 & 32.7 & 40.7 & 36.0 & 210.8 \\
        FSAF~\cite{zhu2019feature} & ResNet-101 & 23.9 & 41.8 & 23.6 & 14.7 & 34.1 & 40.5 & 55.0 & 290.0 \\
        ATSS~\cite{zhang2020bridging} & ResNet-18 & 20.6 & 35.5 & 20.5 & 12.0 & 30.7 & 35.3 & 19.0 & 158.2 \\
        ATSS~\cite{zhang2020bridging} & ResNet-50 & 24.0 & 40.2 & 24.6 & 14.3 & 35.7 & 40.9 & 31.9 & 209.9 \\
        ATSS~\cite{zhang2020bridging} & ResNet-101 & 24.6 & 40.9 & 25.5 & 14.7 & 36.6 & 41.3 & 50.9 & 289.0 \\
        GFL~\cite{li2020generalized} & ResNet-18 & 25.9 & 44.8 & 26.0 & 16.1 & 37.5 & 40.7 & 19.1 & 161.4 \\
        GFL~\cite{li2020generalized} & ResNet-50 & 28.8 & 48.8 & 29.4 & 18.5 & 41.2 & 47.3 & 32.1 & 213.1 \\
        GFL~\cite{li2020generalized} & ResNet-101 & 29.0 & 49.2 & 29.6 & 18.7 & 41.4 & 47.2 & 51.1 &  292.2 \\
        TOOD~\cite{feng2021tood} & ResNet-18 & 22.6 & 37.5 & 23.3 & 13.4 & 33.6 & 42.2 & 18.9 & 136.5 \\
        TOOD~\cite{feng2021tood} & ResNet-50 & 25.2 & 41.2 & 26.2 & 15.3 & 37.3 & 41.9 & 31.8 & 188.1 \\
        TOOD~\cite{feng2021tood} & ResNet-101 & 26.1 & 42.3 & 27.5 & 16.0 & 38.3 & 45.7 & 50.8 & 267.3 \\
        VFL~\cite{zhang2021varifocalnet} & ResNet-18 & 24.1 & 39.9 & 24.9 & 14.6 & 35.1 & 41.0 & 19.6 & 145.3\\
        VFL~\cite{zhang2021varifocalnet} & ResNet-50 & 27.0 & 43.8 & 28.1 & 16.9 & 39.1 & 45.4 & 32.5 & 196.9\\
        VFL~\cite{zhang2021varifocalnet} & ResNet-101 & 27.3 & 44.6 & 28.5 & 17.6 & 39.0 & 44.9 & 51.5 & 276.1\\
        GFLv2~\cite{li2021generalized} & ResNet-18 & 24.3 & 40.0 & 25.5 & 14.6 & 35.9 & 40.5 & 19.1 & 161.5\\
        GFLv2~\cite{li2021generalized} & ResNet-50 & 26.6 & 43.0 & 28.1 & 16.4 & 39.0 & 44.4 & 32.1 & 213.1\\
        GFLv2~\cite{li2021generalized} & ResNet-101 & 26.9 & 43.8 & 28.3 & 16.8 & 39.5 & 45.5 & 51.1 & 292.2\\
        QueryDet$^{\dag}$~\cite{yang2022querydet} & ResNet-50 & 28.3 & 48.1 & 28.8 & - & - & - & - & - \\
        CEASC~\cite{du2023adaptive} & ResNet-18 & 26.0 & 44.5 & 26.4 & 16.8 & 37.8 & 43.3 & 19.3 & - \\
        CEASC~\cite{du2023adaptive} & ResNet-50 & 28.9 & 48.6 & 29.5 & 19.3 & 41.4 & 44.0 & 32.2 & - \\
        CEASC~\cite{du2023adaptive} & ResNet-101 & 29.1 & 49.4 & 29.5 & 18.8 & 41.3 & 47.7 & 51.2 & - \\
        \midrule
        GFL + CFPT (ours) & ResNet-18 & 26.7 & 46.1 & 26.8 & 17.0 & 38.4 & 42.2 & 19.4 & 163.2\\
        GFL + CFPT (ours) & ResNet-50 & 29.5 & 49.7 & 30.1 & 19.6 & 41.5 & 47.1 & 32.3 & 214.8\\
        GFL + CFPT (ours) & ResNet-101 & \textbf{29.7} & \textbf{50.0} & \textbf{30.4} & \textbf{19.7} & \textbf{41.9} & \textbf{48.0} & 51.3 & 293.9 \\
        \bottomrule
      \end{tabular}}
    \end{table*}

\section{Experiments}
\subsection{Datasets}
We evaluate the effectiveness of the proposed CFPT on three challenging datasets specifically designed for small object detection from the drone's perspective: VisDrone2019-DET~\cite{du2019visdrone}, TinyPerson~\cite{yu2020scale}, and xView~\cite{li2020object}.

\subsubsection{VisDrone2019-DET}
This dataset comprises 7,019 images captured by drones, with 6,471 images allocated for training and 548 images for validation. It includes annotations for ten categories: bicycle, awning tricycle, tricycle, van, bus, truck, motor, pedestrian, person, and car. The images have an approximate resolution of $2000 \times 1500$ pixels.

\subsubsection{TinyPerson}
This dataset is collected by drones and is mainly used for small object detection in long-distance scenarios, where the objects have an average length of less than 20 pixels. It contains 1,610 images, with 794 allocated for training and 816 for testing. The dataset comprises 72,651 labeled instances categorized into two groups: ``sea person" and ``earth person". For simplicity, we merge the above two categories into a single category named ``person".

\subsubsection{xView}
This dataset is one of the largest publicly available high-altitude image collections, consisting of WorldView-3 satellite imagery with a 0.3 $m$ ground sampling distance. Spanning over 1,400 $km^{2}$, it includes more than 1 million objects across 60 categories, all annotated with bounding boxes. It features diverse global scenes, with an emphasis on small, rare, fine-grained, and multi-type objects, and offers a resolution that surpasses most public satellite datasets. Following the settings of CtxMIM~\cite{ctxmim} and MTP~\cite{wang2024mtp}, we randomly select 700 images for training and 146 images for testing.

\subsection{Implementation Details}
We implement the proposed CFPT using PyTorch~\cite{paszke2019pytorch} and the MMdetection Toolbox~\cite{chen2019mmdetection}. All experiments on the VisDrone2019-DET and TinyPerson datasets are conducted using a single RTX 3090 GPU with a batch size of 2. We use SGD as the optimizer for model training with a learning rate of 0.0025, momentum of 0.9, and weight decay of 0.0001. We conduct ablation studies and compare the performance of various state-of-the-art feature pyramid networks on the VisDrone2019-DET dataset, using an input resolution of $1333 \times 800$ and 1$\times$ training schedule (12 epochs). To accelerate the convergence of the model, the linear warmup strategy is employed at the beginning of training. For performance comparisons among various state-of-the-art detectors on the VisDrone2019-DET dataset, we train the models for 15 epochs to ensure full convergence following CEASC~\cite{du2023adaptive}. 

In our experiments on the TinyPerson dataset~\cite{yu2020scale}, we mitigate excessive memory usage by dividing the high-resolution images into evenly sized blocks with a $30\%$ overlap ratio. Each block is scaled proportionally, ensuring the shortest side measures 512 pixels. For model training and testing, we use a batch size of 1 and a 1$\times$ schedule, incorporating both multi-scale training and multi-scale testing to provide a comprehensive evaluation of the model's performance.

For comparisons on the xView dataset, we use four RTX 3090 GPUs, each configured with a batch size of 2, while maintaining consistency with the hyperparameter settings of MTP~\cite{wang2024mtp} to ensure a fair evaluation.

\subsection{Comparison with Other Feature Pyramid Networks}
We begin by comparing the performance of the proposed CFPT with various state-of-the-art feature pyramid networks based on RetinaNet~\cite{lin2017focal} on the VisDrone2019-DET dataset. As shown in Table~\ref{tab:vari_fpn}, our CFPT achieves the best results across different backbone networks, including ResNet-18, ResNet-50, and ResNet-101, achieving the best trade-off between performance and computational complexity. Additionally, compared to SSFPN, which is designed for small object detection in aerial images, our CFPT achieves superior performance (+0.8 AP, +0.5 AP, and +0.4 AP) with fewer parameters (-3.5 M) and lower FLOPs (-55.5 G). 

Furthermore, even with an extended training schedule, such as the 3$\times$ schedule (36 epochs), the proposed CFPT consistently achieves superior performance and computational efficiency on RetinaNet with ResNet-18 and ResNet-50. It surpasses the second-best method, SSFPN, by +0.4 AP and +0.7 AP, respectively, while significantly reducing computational complexity by 55.5 GFLOPs and model size by 3.5 M parameters. These results demonstrate the effectiveness of CFPT in improving small object detection in aerial images.

\begin{table*}[t]
\centering
  \caption{Performance comparison with other state-of-the-art detectors on the TinyPerson dataset. The best performance is highlighted in bold.}
\renewcommand{\arraystretch}{1.1}
  \label{tab:vari_det_tiny}
  \resizebox{\linewidth}{!}{
  \begin{tabular}{l|ccccccc|cc}
    \toprule
    Method & AP-tiny(\%) & AP-tiny$_{25}$(\%) & AP-tiny1$_{25}$(\%) & AP-tiny2$_{25}$(\%) & AP-tiny3$_{25}$(\%) & AP-tiny$_{50}$(\%) & AP-tiny$_{75}$(\%) & Params(M) & FLOPs(G)\\
    \midrule
    RetinaNet & 29.9 & 59.4 & 33.1 & 61.3 & 79.1 & 27.9 & 2.4 & 36.1 & 212.5 \\
    PISA & 33.2 & 56.7 & 58.9 & 59.1 & 80.8 & 37.5 & 5.4 & 41.0 & 210.3\\
    Reppoints & 33.4 & 66.6 & 45.6 & 70.8 & 78.8 & 31.7 & 2.0 & 36.6 & 197.3\\
    LibraRCNN & 35.6 & 60.9 & 57.7 & 59.1 & 82.9 & 41.2 & 4.8 & 41.2 & 211.4\\
    FasterRCNN & 36.5 & 64.2 & 56.2 & 64.7 & 84.7  & 41.0 & 4.2 & 41.0 & 210.3\\
    DynamicRCNN & 36.8 & 63.8 & 54.0 & 61.7 & 80.4 & 41.5 & 5.0 & 41.0 & 210.3 \\
    PAA & 37.9 & 66.8 & 47.8 & 68.1 & 79.7 & 41.8 & 5.0 & 31.9 & 209.4\\
    CascadeRCNN & 38.8 & 66.9 & 51.0 & 67.6 & 84.5 & 44.4 & 5.2 & 68.9 & 242.2 \\
    VFL & 39.0 & 68.5 & 54.6 & 73.4 & 79.7 & 43.0 & 5.4 & 32.5 & 196.5 \\
    GFL & 41.8 & 70.9 & 54.5 & 75.0 & 82.2 & 48.1 & 6.6 & 32.0 & 212.6 \\
    FSAF & 42.5 & 74.3 & 62.1 & 76.2 & 83.3 & 47.5 & 5.6 & 36.0 & 210.4 \\
    \midrule
    GFL + CFPT (ours) & 44.2 & 74.6 & \textbf{62.5}  & 77.8 & 83.6 & 51.0 & \textbf{7.0} & 32.3 & 214.3\\
    FSAF + CFPT (ours) & \textbf{44.5} & \textbf{76.4} & 62.3 & \textbf{79.2}  & \textbf{85.4} & \textbf{51.4} & 5.8 & 37.0 & 212.3\\
    \bottomrule
  \end{tabular}}
\end{table*}

\subsection{Comparison with State-of-the-Art Methods}
To further evaluate the effectiveness of CFPT, we replace the feature pyramid network in state-of-the-art detectors with CFPT and compare their performance on the VisDrone2019-DET, TinyPerson, and xView datasets.

\subsubsection{VisDrone2019-DET} We replace the FPN in GFL~\cite{li2020generalized} with CFPT and compare its performance with various state-of-the-art detectors. As shown in Table~\ref{tab:vari_det}, integrating CFPT into GFL improves its performance by 0.8 AP, 0.7 AP, and 0.7 AP on ResNet-18, ResNet-50, and ResNet-101, respectively, with only a slight increase in the number of parameters by 0.3 M, 0.2 M, and 0.2 M. Compared to CEASC~\cite{du2023adaptive}, our method increases the number of parameters by only 0.1 M, yet achieves notable performance improvements (+0.7 AP, +0.6 AP, and +0.6 AP), demonstrating the effectiveness of CFPT.

\subsubsection{TinyPerson} For the comparison on the TinyPerson dataset, we use the evaluation metrics defined in ~\cite{yu2020scale} to thoroughly assess the model performance. We observe that GFL~\cite{li2020generalized} excels in fine-grained detection, as indicated by its superior performance on the AP-tiny$_{75}$ metric, while FSAF~\cite{zhu2019feature} demonstrates greater effectiveness for coarse-grained prediction, as reflected in its better performance on the AP-tiny$_{25}$ and AP-tiny$_{50}$ metrics. Therefore, we integrate CFPT into both GFL and FSAF to evaluate its adaptability to these two scenarios. As shown in Table~\ref{tab:vari_det_tiny}, CFPT delivers significant performance improvements, including a 2.4 AP-tiny gain for GFL (44.2 AP-tiny v.s 41.8 AP-tiny) and a 2.0 AP-tiny gain for FSAF (44.5 AP-tiny v.s 42.5 AP-tiny), with all performance metrics showing enhancements. Therefore, integrating CFPT effectively improves the small object detection performance of the model, demonstrating its efficacy for small object detection in aerial images.

\subsubsection{xView} For the comparison on the xView dataset, we adopt the latest MTP~\cite{wang2024mtp} as the baseline model, which has demonstrated competitive performance in remote sensing object detection tasks. To assess the impact of our proposed CFPT, we replace the traditional FPN with CFPT in the MTP architecture. As illustrated in Table~\ref{tab:vari_det_xview}, the integration of CFPT into MTP delivers a significant performance improvement of 2.3 AP$_{0.5}$ (21.7 AP$_{0.5}$ v.s 19.4 AP$_{0.5}$), demonstrating the effectiveness of CFPT in enhancing small object detection.

\begin{table}[t]
\centering
  \caption{Performance comparison with other state-of-the-art detectors on the xView dataset. The best performance is highlighted in bold. The ``IN1K" represents supervised learning on ImageNet-1K, while ``IN22K" denotes pretraining on ImageNet-22K.}
\renewcommand{\arraystretch}{1.1}
  \label{tab:vari_det_xview}
  \resizebox{\linewidth}{!}{
  \begin{tabular}{l|c|c}
    \toprule
    Method & Backbone & AP$_{0.5}$(\%)\\
    \midrule
      RetinaNet~\cite{lin2017focal} &  ResNet-50  & 10.8 \\
      IN1K & ResNet-50  & 14.4\\
      IN1K & Swin-B   & 16.3 \\
      GASSL \cite{gassl} & ResNet-50 & 17.7 \\
      SeCo \cite{seco} &  ResNet-50 & 17.2 \\
      CACO \cite{caco} &  ResNet-50  & 17.2 \\
      CtxMIM \cite{ctxmim} & Swin-B  & 18.8 \\
      MAE~\cite{rvsa} & ViT-B + RVSA & 14.6 \\
      MAE~\cite{rvsa} & ViT-L + RVSA & 15.0 \\
      IN22K & InternImage-XL & 17.0 \\
      MAE + MTP~\cite{wang2024mtp} & ViT-L + RVSA & 19.4 \\
      \midrule
      MAE + MTP + CFPT (ours) & ViT-L + RVSA & \textbf{21.7} \\ 
    \bottomrule
  \end{tabular}}
\end{table}

\subsubsection{Inference Speed} 
We evaluate the efficiency of the proposed CFPT by comparing its frames per second (FPS) with several state-of-the-art feature pyramid networks. Specifically, we use the RetinaNet with ResNet-50 as the baseline model and assess both performance and FPS by incorporating competitive feature pyramid networks into this framework. As shown in Table~\ref{tab:vari_inf_speed}, CFPT achieves higher FPS than most feature pyramid networks while also exhibiting superior  AP (e.g., DRFPN, FPT, AFPN). Notably, despite CFPT having lower FLOPs than PAFPN and SSFPN, its inference speed is slightly lower. This discrepancy is attributed to the more efficient optimization of convolutional operations compared to attention mechanisms. However, as FLOPs represent the theoretical upper bound of inference speed, further acceleration can be achieved through customized CUDA operators. We plan to explore this issue in future research to improve the efficiency.

\begin{table}[t]
\centering
  \caption{Comparison of performance and FPS between CFPT and several state-of-the-art feature pyramid networks.}
\renewcommand{\arraystretch}{1.1}
  \label{tab:vari_inf_speed}
  \resizebox{1\linewidth}{!}{
  \begin{tabular}{l|c|ccc}
    \toprule
    Method & Backbone & AP(\%) & FPS(s/img) & FLOPs(G)\\
    \midrule
    RetinaNet~\cite{lin2017focal} & ResNet-50 & 18.1 & \textbf{25.6} & 203.7 \\
    PAFPN~\cite{liu2018path} & ResNet-50 & 21.2 & 22.5 & 222.7 \\
    DRFPN~\cite{ma2020dual} & ResNet-50 & 21.5 & 17.9 & 228.5 \\
    FPT~\cite{zhang2020feature} & ResNet-50 & 19.3 & 11.8 & 331.8 \\
    SSFPN~\cite{hong2021sspnet} & ResNet-50 & 21.7 & 21.0 & 274.0 \\
    AFPN~\cite{yang2023afpn} & ResNet-50 & 20.7 & 13.4 & 250.0 \\
    CFPT & ResNet-50 & \textbf{22.2} & 20.4 & 218.5 \\
    \bottomrule
  \end{tabular}}
\end{table}

\subsection{Ablation Study}
\subsubsection{Order of CCA and CSA}
\begin{figure}[!t]
  \centering
  \includegraphics[width=1\linewidth]{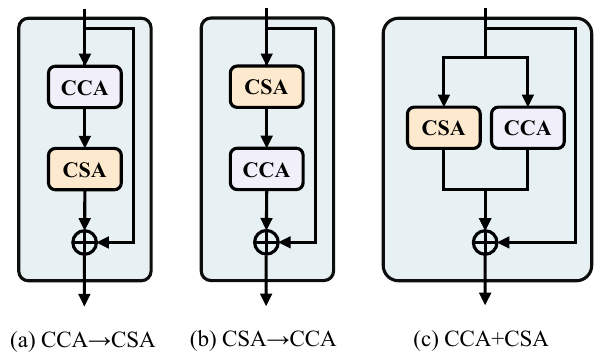}
  \caption{Three usage patterns of CCA and CSA in CAM. Note that the input and output feature maps of CCA and CSA have identical shapes, ensuring the compatibility of these three solutions.}
  \label{fig:vari_cam_mode}
\end{figure}

We investigate the impact of the application order of CCA and CSA on model performance. Specifically, we compare the performance of three configurations, as illustrated in Fig.~\ref{fig:vari_cam_mode}, including CCA$\rightarrow$CSA, CSA$\rightarrow$CCA, and CCA$+$CSA. As reported in Table~\ref{tab:vari_cam_mode}, the CCA$\rightarrow $CSA configuration achieves optimal performance, attaining an AP of 22.2. This result can be attributed to the complementary roles of CCA and CSA. When CCA is applied first, it provides a global receptive field along the spatial dimension, allowing CSA to leverage this global contextual information to generate more precise attention maps and capture finer neighboring details. Conversely, applying CSA before CCA may disrupt the locality of spatial features due to CSA's global receptive field along the channel dimension. This disruption can hinder CCA from effectively focusing on spatial-wise neighboring information. Additionally, the CCA$+$CSA configuration prevents interaction between these two modules, thereby limiting their ability to utilize each other's information for enhanced feature aggregation.

\begin{figure*}[t!]
  \centering
  \includegraphics[width=1\linewidth]{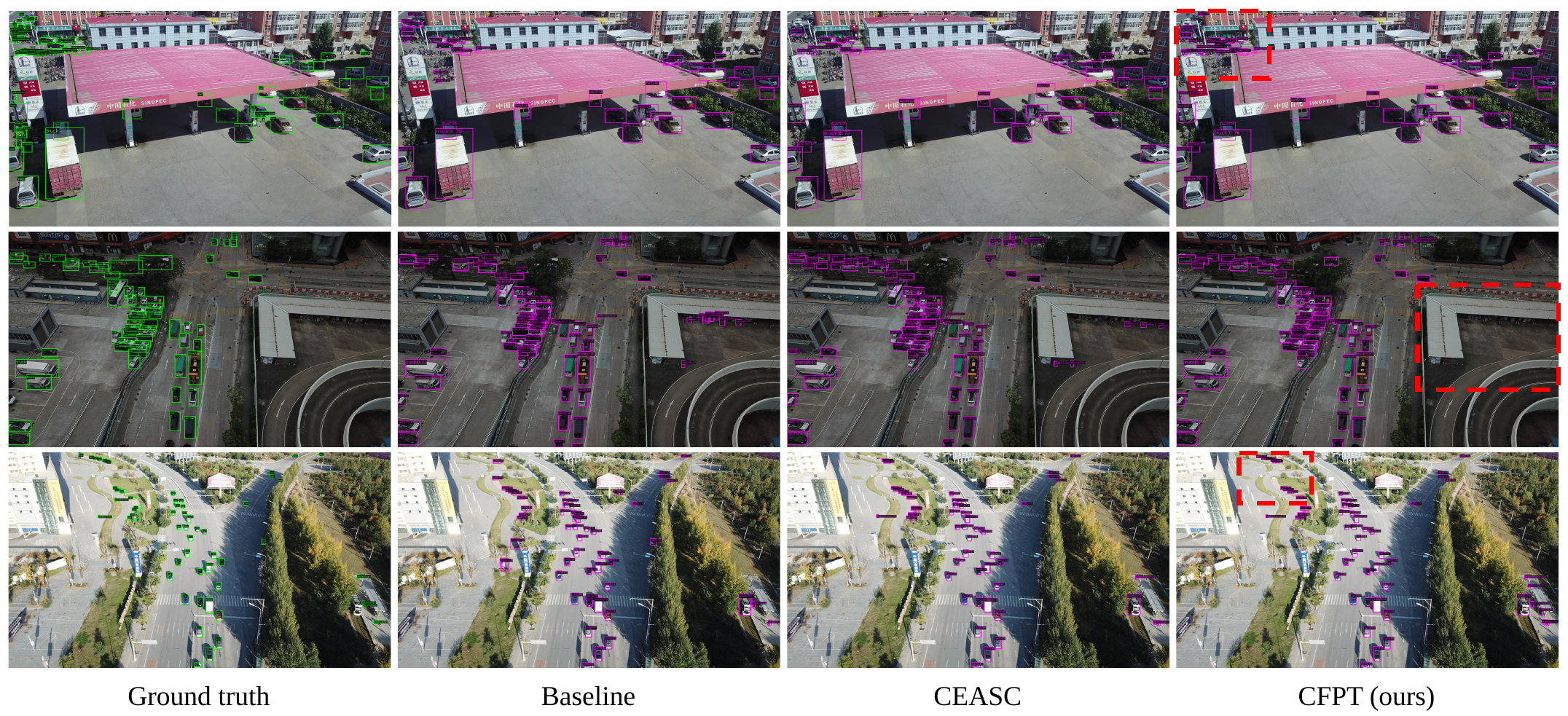}
  \caption{Visualization of detection results of baseline model (\ie GFL), CEASC, and CFPT on the VisDrone2019-DET dataset. The boxes with red dashed lines highlight areas with significant differences.}
  \label{fig:visual_on_visdrone}
\end{figure*}

\begin{table}[!t]
  \footnotesize
  \centering
  \renewcommand\arraystretch{1.2}
  \setlength{\tabcolsep}{1.2mm}
  \caption{Performance impact of the order of CCA and CSA in CAM.}
  \begin{tabular}{c|cccc}
    \toprule
    Method & AP(\%)& AP$_{0.5}$(\%) & AP$_{0.75}$(\%) & AP-small(\%)\\
    \midrule
    CCA $\rightarrow$  CSA   & \textbf{22.2} & \textbf{38.0} & \textbf{22.4} & \textbf{11.9} \\
    CSA $\rightarrow$  CCA   & 22.1 & 37.7 & \textbf{22.4} & \textbf{11.9} \\
    CCA $+$ CSA   & 22.0 & 37.4 & 22.2 & 11.6 \\
    \bottomrule
    \end{tabular}
    \label{tab:vari_cam_mode}
    \vspace{-8pt}
\end{table}

\begin{figure}[!t]
  \centering
  \includegraphics[width=1\linewidth]{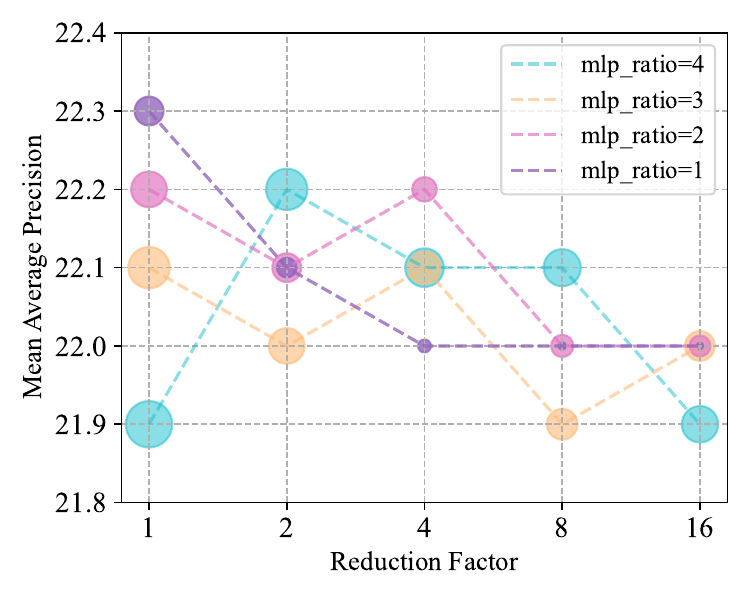}
  \caption{The impact of channel size reduction factor and MLP ratio on model performance. The size of each bubble represents the FLOPs of the corresponding model.}
  \label{fig:reduction_factor_mlp_ratio}
\end{figure}

\subsubsection{Effectiveness of each proposed component}
We evaluate the effectiveness of each component by progressively integrating the proposed modules into the baseline model (i.e., RetinaNet without FPN).
As indicated in Table~\ref{tab:grad_applied}, the integration of CCA and CSA significantly improves the baseline model, with increases of +3.4 AP and +3.2 AP, respectively. The combination of CCA and CSA further enhances performance, resulting in a notable AP increase of +3.9 (22.0 AP vs. 18.1 AP). The subsequent incorporation of CCPE provides additional improvements, yielding an AP gain of +0.1 for CCA and +0.2 for CSA. When all modules are combined into a complete CAM, our method achieves an overall AP of 22.2, surpassing the baseline model by a substantial margin of +4.1 AP (22.2 AP v.s 18.1 AP). Notably, integrating either CCA or CSA alone outperforms most feature pyramid networks presented in Table~\ref{tab:vari_fpn}, highlighting their potential for small object detection in aerial images.

We also report the impact of each proposed component on the model's computational complexity, number of parameters, and inference speed in Table~\ref{tab:grad_applied}. Utilizing only a single component (\eg CCA), CFPT introduces an additional 1.4 M parameters, 7.4 G FLOPs, and 0.004 s/img of inference latency compared to the baseline model, while achieving significant performance gains (+3.5 AP). When all components are integrated, CFPT introduces an additional 2.8 M parameters, 14.8 G FLOPs, and 0.01 s/img of inference latency, while achieving significant performance gains (+4.1 AP). Therefore, CFPT can achieve a better balance between performance and computational complexity.

\begin{figure*}[t!]
  \centering
  \includegraphics[width=1\linewidth]{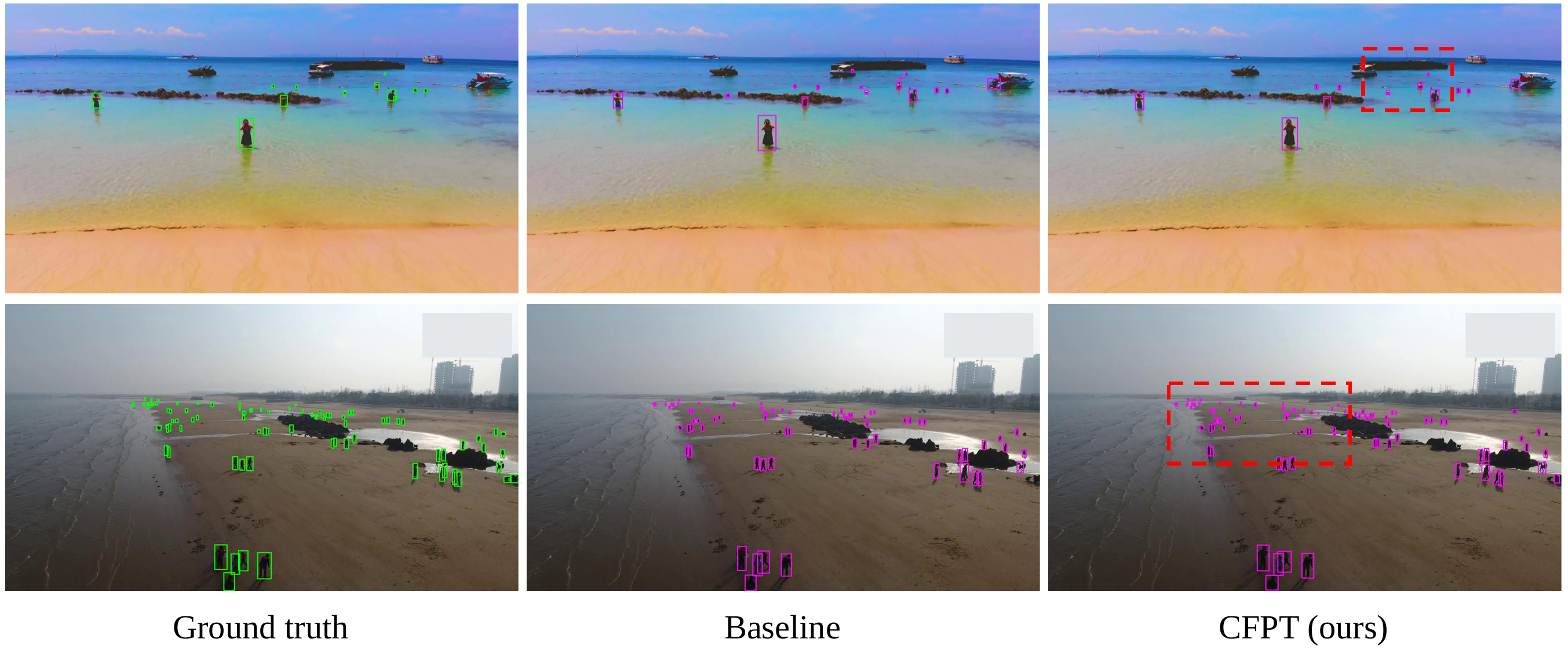}
  \caption{Visualization of detection results of the baseline model (\ie FSAF) and CFPT on TinyPerson dataset. The boxes with red dashed lines highlight areas with significant differences.}
  \label{fig:visual_on_tinyperson}
\end{figure*}

\begin{table*}[t!]
  \footnotesize
  \centering
  \caption{Ablation study of the proposed modules.}
  \renewcommand\arraystretch{1.2}
  \setlength{\tabcolsep}{1.2mm}
      \begin{tabular}{ccc|cccc|ccc}
      \toprule
      CCA & CSA & CCPE & AP(\%)& AP$_{0.5}$(\%) & AP$_{0.75}$(\%) & AP-small(\%) & Params(M) & FLOPs(G) & Inference Speed(s/img) \\
      \midrule
      \multicolumn{3}{c|}{Baseline~\cite{lin2017focal}} & 18.1 & 31.1 & 18.3 & 8.8 & 34.5 & 203.7 & 0.039 \\
      \midrule
      $\checkmark$ & $\times$ & $\times$ & 21.5 & 36.9 & 21.8 & 11.1 & 35.9 & 211.1 & 0.043 \\
      $\times$ & $\checkmark$ & $\times$ & 21.3 & 37.0 & 21.4 & 11.4 & 35.9 & 211.1 & 0.044 \\
      $\checkmark$ & $\times$ & $\checkmark$ & 21.6 & 36.8 & 21.9 & 11.2 & 35.9 & 211.1 & 0.043 \\
      $\times$ & $\checkmark$ & $\checkmark$ & 21.5 & 36.8 & 21.7 & 11.5 & 35.9 & 211.1 & 0.044 \\
      $\checkmark$ & $\checkmark$ & $\times$ & 22.0 & 37.6 & 22.2 & 11.8 & 37.2 & 218.5 & 0.049 \\  
      $\checkmark$ & $\checkmark$ & $\checkmark$ & \textbf{22.2(+4.1)} & \textbf{38.0} & \textbf{22.4} & \textbf{11.9} & 37.3 & 218.5 & 0.049 \\ 
      \bottomrule
      \end{tabular}
      \label{tab:grad_applied}
      \vspace{-8pt}
\end{table*}

\subsubsection{Number of CAMs}
We assess the impact of the number of CAMs on model performance. As shown in Table~\ref{tab:stk_number}, increasing the number of CAMs consistently enhances the model's performance. With three CAMs, the model achieves an AP of 22.5, marking a gain of 4.4 AP compared to the baseline (22.5 AP v.s 18.1 AP). To better balance computational complexity and performance, we set the stacking number of CAM to 1 in all other experiments, even though more CAMs would bring more benefits.

\begin{table}[t!]
  \footnotesize
  \centering
  \renewcommand\arraystretch{1.2}
  \setlength{\tabcolsep}{1.2mm}
  \caption{Performance impact of the stacking number of CAMs.}
  \begin{tabular}{c|ccccc}
    \toprule
    Number & AP(\%)& AP$_{0.5}$(\%) & AP$_{0.75}$(\%) & AP-small(\%) & Params(M)\\
    \midrule
    1  & 22.2 & 38.0 & 22.4 & 11.9 & 37.3\\
    2  & 22.3 & 38.0 & 22.7 & 12.2 & 40.0\\
    3  & \textbf{22.5} & \textbf{38.5} & \textbf{22.8} & \textbf{12.3} & 42.8\\
    \bottomrule
    \end{tabular}
    \label{tab:stk_number}
    \vspace{-8pt}
\end{table}

\subsubsection{Channel Size reduction factor and MLP ratio}
We investigate the effects of various channel size reduction factors (i.e., the compression factor of feature map channels for attention interactions) and MLP ratios (i.e., the expansion factor for channel size in FFNs), aiming to identify the optimal combination that balances computational complexity and model performance. As shown in Fig.~\ref{fig:reduction_factor_mlp_ratio}, the model achieves the optimal balance between computational complexity and performance when the channel size reduction factor is set to 4 and the MLP ratio is set to 2. Therefore, for all experiments conducted on the VisDrone2019-DET and TinyPerson datasets, we consistently use this combination scheme.

\subsubsection{Spatial-wise and channel-wise expansion factors}
Finally, we evaluate the influence of spatial-wise and channel-wise expansion factors on model performance. As demonstrated in Table~\ref{tab:sc_factor}, the model achieves the highest performance when both $\alpha$ and $\beta$ are set to 3. However, increasing $\alpha$ and $\beta$ beyond this value leads to higher computational costs with only marginal improvements in effectiveness. To achieve a better balance between computational efficiency and performance, we set both $\alpha$ and $\beta$ to 1.

\begin{table}[t!]
  \footnotesize
  \centering
  \renewcommand\arraystretch{1.2}
  \setlength{\tabcolsep}{2mm}
  \caption{Performance impact of the spatial-wise and channel-wise expansion factors.}
  \begin{tabular}{cc|ccccc}
    \toprule
    $\alpha$ & $\beta$ & AP(\%)& AP$_{0.5}$(\%) & AP$_{0.75}$(\%) & AP-small(\%) & FLOPs(G)\\
    \midrule
    1 & 1 & 22.2 & 38.0 & 22.4 & \textbf{11.9} & 218.5\\
    3 & 1 & 22.1 & 37.9 & 22.2 & 11.5 & 218.6\\
    1 & 3 & 22.2 & 38.3 & 22.7 & 11.8 & 218.9\\
    3 & 3 & \textbf{22.4} & \textbf{38.5} & \textbf{22.8} & \textbf{11.9} & 219.0\\
    5 & 5 & \textbf{22.4} & 38.3 & 22.5 & 11.7 & 220.3\\
    7 & 7 & \textbf{22.4} & 38.4 & 22.7 & 11.7 & 222.1\\
    \bottomrule
    \end{tabular}
    \label{tab:sc_factor}
    \vspace{-8pt}
\end{table}

\begin{figure*}[t!]
  \centering
  \includegraphics[width=1\linewidth]{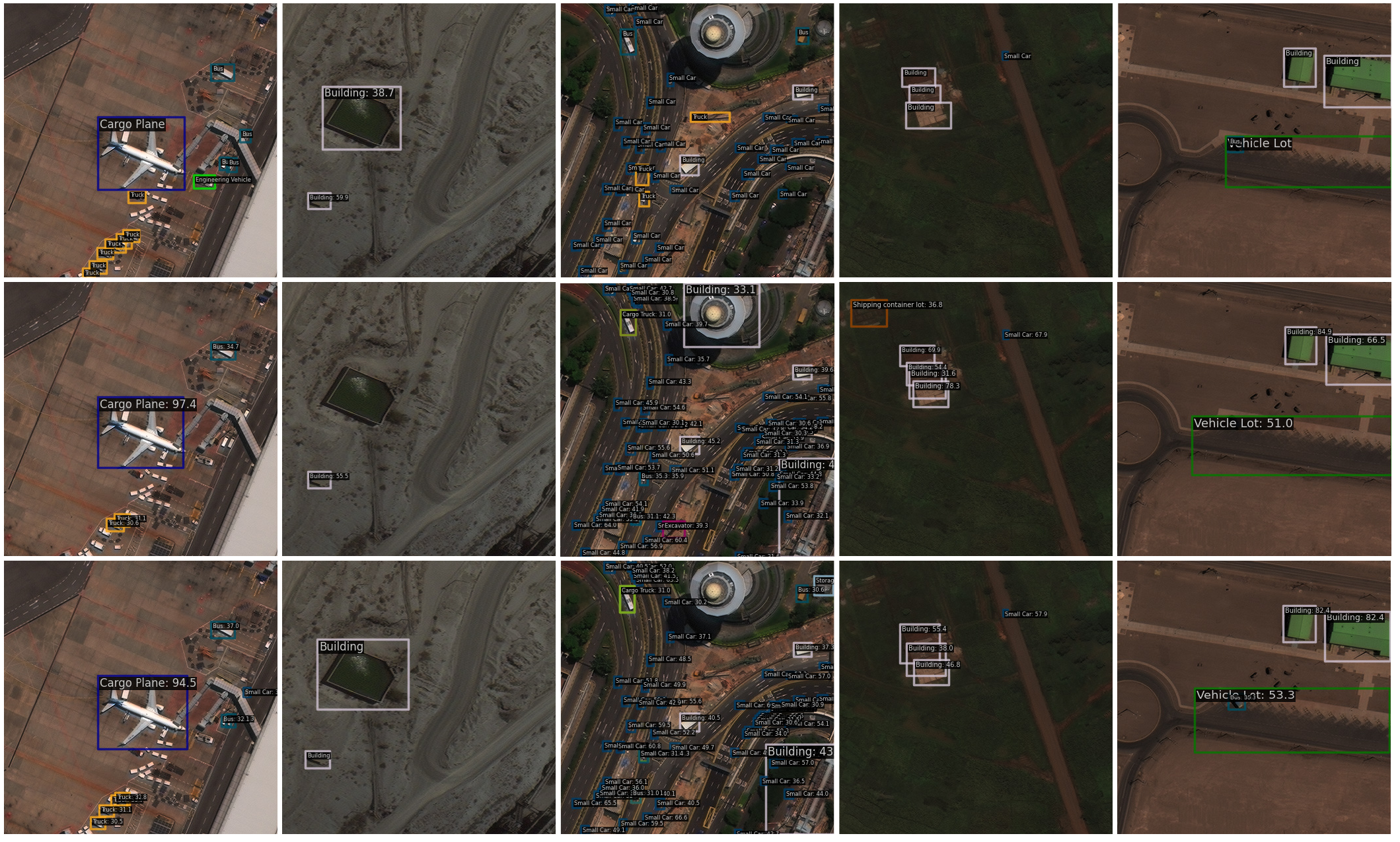}
  \caption{Visualization of detection results of the baseline model (\ie MAE + MTP) and CFPT (\ie MAE + MTP + CFPT) on the xView dataset. The first row shows the ground truths, the second row shows the results for the baseline model, and the third row shows the results for CFPT.}
  \label{fig:visual_on_xview}
\end{figure*}

\subsection{Qualitative Analysis}
We conduct the qualitative analysis of CFPT by visualizing the detection results on the VisDrone2019-DET, TinyPerson, and xView datasets, with a confidence threshold of 0.3 for all visualizations. As shown in Fig.~\ref{fig:visual_on_visdrone}, we apply CFPT to GFL and compare the detection results of (\ie GFL), CEASC, and CFPT on the VisDrone2019-DET dataset. The application of CFPT effectively reduces both the missed detection rate (first and third rows) and the false detection rate (second row), thereby enhancing the model's overall performance. In addition, the third row in Fig.~\ref{fig:visual_on_visdrone} demonstrates CFPT's effectiveness in detecting small objects. Furthermore, the detection results on the TinyPerson and xView datasets, shown in Fig.~\ref{fig:visual_on_tinyperson} and Fig.~\ref{fig:visual_on_xview}, further support the above claims, demonstrating that CFPT effectively reduces both missed and false detection rates while improving the model's ability to detect small objects.

\section{Conclusion}
This paper introduces CFPT, a novel upsampler-free feature pyramid network for small object detection in aerial images. CFPT can explicitly focus more on shallow feature maps and eliminate the static kernel-based interaction scheme to mitigate the impact of scale variations on model performance, making it particularly well-suited for object detection in aerial images. Specifically, CFPT consists of two meticulously designed attention blocks with linear computational complexity: CCA and CSA. These two modules capture contextual information from different perspectives, and their integration equips the model with crucial global contextual modeling capability for small object detection. To improve positional awareness during cross-layer interactions, we propose a new positional encoding method called CCPE. Extensive experiments on three challenging aerial datasets demonstrate that CFPT outperforms state-of-the-art feature pyramid networks while reducing computational costs. In future work, we plan to explore deformable cross-layer interaction solutions and investigate more efficient implementation strategies.


\bibliographystyle{IEEEtran}
\bibliography{main}

\end{document}